\documentclass[a4paper,fleqn]{cas-sc}

\usepackage[numbers,square]{natbib}

\usepackage{placeins}

\def\tsc#1{\csdef{#1}{\textsc{\lowercase{#1}}\xspace}}
\tsc{WGM}
\tsc{QE}

\begin{document}
\let\WriteBookmarks\relax
\def\floatpagepagefraction{1}
\def\textpagefraction{.001}

\shorttitle{}    

\shortauthors{M. Antal et~al.}

\title [mode = title]{TRAIL-UKB: A sparse-autoencoder atlas of disease-trajectory transformer representations}  

\author[1]{Mátyás Antal}

\author[1]{Nara Csutora}

\author[1]{Márk Marosi}

\author[2]{Tamás Nagy}

\author[2,3,4]{Gabriella Juhász}

\author[1]{Péter Antal}

\author[1]{András Gézsi}

\affiliation[1]{
  organization={Department of Artificial Intelligence and Systems Engineering, Faculty of Electrical Engineering and Informatics, Budapest University of Technology and Economics},
  city={Budapest},
  country={Hungary}
}

\affiliation[2]{
  organization={Department of Pharmacodynamics, Faculty of Pharmaceutical Sciences, Semmelweis University},
  addressline={Nagyvárad tér 4},
  city={Budapest},
  postcode={1089},
  country={Hungary}
}

\affiliation[3]{
  organization={Center of Pharmacology and Drug Research \& Development, Semmelweis University},
  addressline={Üllői Street 26},
  city={Budapest},
  postcode={1085},
  country={Hungary}
}

\affiliation[4]{
  organization={NAP3.0-SE Neuropsychopharmacology Research Group, Hungarian Brain Research Program, Semmelweis University},
  addressline={Üllői Street 26},
  city={Budapest},
  postcode={1085},
  country={Hungary}
}

\begin{abstract}
We present TRAIL-UKB, a sparse-autoencoder (SAE) feature atlas of a Delphi-2M-inspired masked-event transformer trained on 5.1 million first-occurrence ICD-10 events from 502,383 UK Biobank participants. SAEs decompose the model's internal representations into sparse, individually interpretable features. An automated pipeline produces feature cards combining clinical context, temporal patterns, genotype associations, model-output effects, logical activation rules and natural-language descriptions.

A feature's activation may reflect a diagnosis, its surrounding context, or both. To distinguish additional contextual selectivity from patterns inherited from the diagnoses themselves, we compare observed feature context with diagnosis-conditioned expectations, weighting each diagnosis's usual context by its share of feature activation mass.

The atlas contains 7,712 features representing diagnosis indicators, comorbidity and temporal contexts, patient strata, genotype-linked enrichments and recording artifacts. SAE reconstructions explain over 99.7\% of hidden-state variance and preserve the model's top-ranked predictions at over 98\% of evaluated positions. Across three independently trained SAE fits, 53\% of supported feature instances recur in at least two. Counterfactuals test feature activation sensitivity, while decoder-direction steering evaluates effects on model predictions. Comparisons with alternative sparse decompositions assess representation quality. Finally, retrospective, prescription-derived treatment-resistant depression classification demonstrates how downstream model scores can be connected to interpretable SAE features through decoder folding and dependence-aware analysis. We release the analysis pipeline and privacy-screened feature cards for selected TRD-associated features.
\end{abstract}

\begin{graphicalabstract}
\includegraphics[width=1\linewidth]{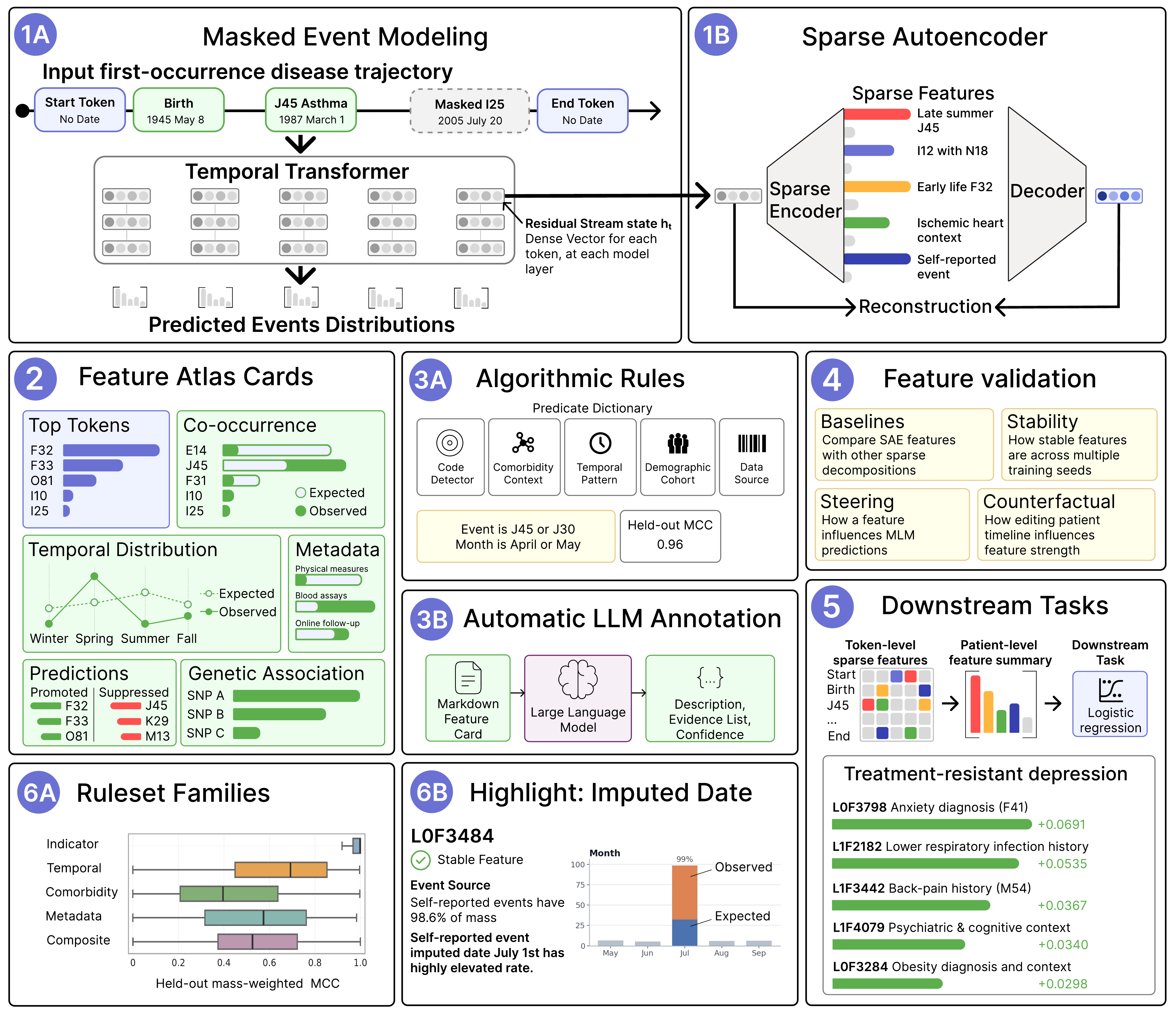}
\label{fig:overview_fig1}
\end{graphicalabstract}


\begin{highlights}
\item Sparse autoencoders expose clinical structure in longitudinal EHR representations.
\item Diagnosis-conditioned baselines distinguish excess context from token composition.
\item Feature cards reveal clinical patterns and recording artifacts across 7,712 features.
\item Decoder folding connects TRD classification to interpretable atlas features.
\end{highlights}


\maketitle

\section{Introduction}\label{introduction}

Transformer models trained on longitudinal health records support disease prediction and the modelling of patient trajectories~\cite{liBEHRTTransformerElectronic2020,rasmy2021medbert,shmatko2025delphi}. Their internal representations may combine disease history with demographic composition, recording practices and other properties of the training cohort. For medical model development, these distinctions matter: a representation associated with a diagnosis may capture an informative comorbidity, an administrative convention, or both. Predictive performance alone does not resolve which explanation applies. Methods that connect internal representations to interpretable clinical and recording context could help researchers assess proposed phenotypes and identify model behaviour requiring further investigation before transfer to another clinical setting.
Our aim in this paper is to help researchers inspect the information represented inside a model trained on recorded diagnosis histories. We assess whether patterns in these representations reflect particular diagnoses, additional context such as age or other diagnoses, recording practices, or a combination of these.

For each event in a patient’s history, the transformer constructs a numerical representation that can incorporate information from elsewhere in that history. We use sparse autoencoders (SAEs) to approximate each representation as a combination of a small number of active components drawn from a larger learned set~\cite{bricken2023monosemanticity,cunningham2023sparse,templeton2024scaling}. We call these components features: they are not predefined clinical variables, and their meaning must be assessed from evidence. The feature atlas is a catalogue that brings together these features, their proposed interpretations and the supporting evidence.

Diagnosis trajectories create a specific problem for interpreting these features. Each diagnosis token already has a characteristic distribution of age, event dates, comorbidities and other patient attributes. A feature may therefore appear context-selective simply because it activates on diagnoses that carry that context. We address this by contrasting observed feature context with an expectation built from the usual context of each activating diagnosis, weighted by its share of feature activation mass. For example, a reflux-associated feature enriched for gastritis should be compared with the gastritis context expected from its activating diagnoses, not only with the overall cohort. Such comparisons distinguish additional selectivity from diagnosis composition while leaving room for residual confounding and incomplete records.

We introduce TRAIL-UKB (Trajectory Representation Atlas for Interpretable Longitudinal features on the UK Biobank), applying this approach to 5.1 million first-occurrence diagnoses from 502,383 participants~\cite{sudlow2015uk,bycroft2018uk}. The inspected model, Delphi-MLM, is a compact bidirectional masked-event transformer inspired by Delphi-2M~\cite{shmatko2025delphi}. We make three contributions:

\begin{enumerate}

\item \textbf{Identifying context beyond what is expected from the diagnoses.} We compare the age, timing and other diagnoses associated with each feature against the usual patterns for the diagnoses on which it activates. This assesses whether the feature emphasizes additional context beyond what would already be expected from those diagnoses in the cohort.

\item \textbf{Building and evaluating an atlas of model features.} We assemble a catalogue describing the clinical and recording-related patterns associated with the features based on the diagnosis trajectories in UK Biobank. We assess whether the decomposition preserves the model’s internal representations and predictions, whether simple rules describe feature activation, and whether the features reappear when the same trained model is decomposed again. We also compare alternative representations and examine how selected features influence predictions.

\item \textbf{Linking a classification score to the feature atlas.} For a retrospective treatment-resistant depression (TRD) classification task with labels derived from prescription records, we rewrite the fitted classifier’s score in terms of the atlas features without changing its value. We consider correlated features together and check that the contributions selected for interpretation remain stable and are not largely offset by other terms in the score.

\end{enumerate}

Together, these contributions provide a method for investigating clinical and recording-related structure in learned representations, and a basis for subsequent circuit analysis (graphical abstract).

\subsection{Related work}
\label{sec:related_work}

Clinical trajectory modelling includes masked-event approaches such as BEHRT and Med-BERT~\cite{liBEHRTTransformerElectronic2020,rasmy2021medbert}, extensions incorporating temporal information~\cite{pang2021cehrbert}, and generative models such as Foresight and Delphi-2M~\cite{kraljevic2024foresight,shmatko2025delphi}. These approaches learn contextual representations for prediction from longitudinal records. Complementary work has characterized multimorbidity using Bayesian dependency maps and temporal models~\cite{marx2017comorbidities,gezsi2024unique}, and combined heterogeneous biobank data in predictive representations~\cite{antal2024modular}. TRAIL-UKB examines the internal features of a trained trajectory model, connecting their behaviour to the clinical histories and recording context they represent.

Input attribution methods, including LIME, integrated gradients and SHAP, relate predictions to observed inputs~\cite{ribeiro2016why,sundararajan2017axiomatic,lundberg2017unified}. Probing classifiers assess information recoverable from internal states, with appropriate controls needed to interpret successful decoding~\cite{alain2016understanding,hewitt2019control,belinkov2022probing}. Sparse dictionary learning offers a complementary decomposition into features that can be examined across many examples~\cite{bricken2023monosemanticity,cunningham2023sparse,gao2024scaling}. Interventions and circuit-discovery methods then investigate whether internal components participate in particular computations~\cite{conmy2023towards,marks2024sparse,lindsey2025biology}. Recoverable information, interpretable feature descriptions and demonstrated computational roles remain distinct evaluation targets.

In medicine, Wu et al.~\cite{wu2024beyond} used dictionary learning to interpret clinical-text models for automated ICD coding and demonstrated effects of feature steering. SAEs have also been applied to protein language models, where features can be compared with structural and functional annotations while also revealing patterns outside existing labels~\cite{simon2024interplm,gujral2025protein}. Our setting differs because the input tokens are already diagnosis codes, and their cohort-specific context supplies a strong alternative explanation for apparent feature meaning. TRAIL-UKB makes this diagnosis-conditioned expectation explicit and combines it with representation comparisons, activation-rule evaluation, retraining and selected interventions. The experimental comparisons assess alternative representations under a common analysis pipeline; they do not establish superior clinical-reader usefulness over attribution methods or simpler clinical summaries.

\section{Materials and methods}

\subsection{Cohort, event representation and participant splits}

We analysed 5.1 million dated first-occurrence diagnoses from 502,383 UK Biobank participants~\cite{sudlow2015uk,bycroft2018uk}. Input sequences contained three-character ICD-10 codes together with birth, death and sequence-boundary tokens (Table~\ref{tab:cohort}). Corresponding event-source fields and demographic, health and assay metadata supported feature annotation and covariate adjustment. General-practice prescriptions supplied the downstream TRD proxy, and genotype annotations used 12,778 variants retained for testing from a selected panel. The field inventory and export procedures are provided in Supplementary Sections~\ref{sec:supp_inventory} and~\ref{sec:supp_preprocessing}.

A fixed participant-level 70/10/20 train/validation/test split was reused across transformer and SAE fitting, ruleset evaluation and downstream modelling. Stratification used supported sex, birth-cohort and outcome-related strata, with rare combinations collapsed before sampling. The final TRD labels were applied to these existing assignments. Model fitting used training participants, model selection used validation participants, and test results were evaluated after selection. Descriptive atlas exports also summarize the full cohort; these summaries are distinguished from held-out evaluations.

\begin{table}[!ht]
\caption{\textbf{Training cohort and diagnosis-sequence dataset.}}
\label{tab:cohort}
\centering
\begin{tabular}{ll}
\hline
Metric & Value \\
\hline
Participants & 502,383 \\
Female / male participants & 273,309 / 229,074 \\
Train / validation / test split & 70\% / 10\% / 20\% \\
Total diagnosis events & 5,136,048 \\
Events per participant, median (IQR) & 8 (4--14) \\
Events per participant, mean (maximum) & 10.2 (115) \\
Event date range & March 1934--March 2020 \\
\hline
\end{tabular}
\end{table}

\subsection{Transformer and sparse-autoencoder representations}

We trained a two-layer bidirectional transformer with hidden dimension 256 to predict masked events. Inputs combined diagnosis and special-token embeddings with token type, ICD-10 block/chapter and calendar year/month/day. Training followed standard masked modelling conventions. We selected 15\% of eligible event positions, including date of birth, and masked, randomly replaced or retained their tokens with probabilities 80\%, 10\% and 10\%, respectively. Calendar information was retained, while chapter and block identifiers were derived from the corrupted token. Cross-entropy was evaluated at the selected positions only. Architecture, training settings and ablations are detailed in Supplementary Section~\ref{sec:supp_mlm_detail}.

We fitted one JumpReLU SAE~\cite{rajamanoharan2024jumping} to clean, non-padding post-block residual states from each frozen transformer layer. Each SAE maps 256 hidden coordinates to 4,096 features. For a hidden vector $h$, define $s(h)=\sqrt{256}/\max(\|h\|_2,10^{-6})$ and normalized state $\widetilde h=s(h)h$. In column-vector notation,

\begin{equation}
u=W_E(\widetilde h-b_D)+b_E,\qquad
z_f=u_f\mathbf{1}[u_f>\exp(\tau_f)],\qquad
\widehat h=(W_Dz+b_D)/s(h).
\end{equation}

Here $W_D$ is the decoder matrix. The decoder reconstructs normalized hidden states; recovering the original scale additionally uses the input state norm. Training combined reconstruction, sparsity and inactive-feature losses, with 200 million token presentations per layer. Layer-specific hyperparameters  are given in Supplementary Section~\ref{sec:supp_sae_detail}.

We evaluated fraction of variance unexplained (FVU), active features per token and feature liveness. Replacing layer outputs with SAE reconstructions measured preservation of model computation through forward KL divergence and top-prediction agreement. We compared the resulting sparse representation to dense hidden states, thresholded hidden and principal-component coordinates, random top-$k$ directions and learned dictionary top-$k$ features (Supplementary Section~\ref{sec:appendix_baseline_construction}).

\subsection{Feature cards and observed-versus-expected context}

Each feature card describes which diagnoses activate the feature, how often and how strongly it activates, and which clinical or recording contexts accompany that activation. Cards also summarize selected genotype associations, activation rules and effects of the feature direction on model outputs. Genotypes and recording-source labels were not inputs to the transformer; here, they provide additional evidence for interpreting its learned features. A positive activation $a_{ijf}$ identifies an active feature $f$ at position $j$ in participant $i$'s sequence. 
Activation mass is the sum of activation strengths over the positions being summarized, so a mass-weighted percentage is a share of total activation, not a percentage of patients or events.

Source-token baselines distinguish additional contextual selectivity from patterns inherited from the diagnoses on which a feature activates. For example, a reflux feature's gastritis context is compared with the context normally accompanying reflux events, rather than with the whole cohort. For features activating on several tokens, the expectation combines each token's usual context, weighted by its share of feature activation mass.

For categorical axis $z$ and bin $b$, let $B_t^z(b)$ be the fixed cohort distribution for token $t$. The observed and expected feature distributions are
\begin{equation}
\begin{aligned}
P_f^z(b)&=\frac{\sum_{i,j}a_{ijf}\mathbf{1}[z_{ij}=b]}{\sum_{i,j}a_{ijf}},\\
\pi_{ft}&=\frac{\sum_{i,j}a_{ijf}\mathbf{1}[x_{ij}=t]}{\sum_{i,j}a_{ijf}},\qquad
Q_f^z(b)=\sum_t\pi_{ft}B_t^z(b).
\end{aligned}
\end{equation}
Token baselines were computed from the full event export independently of SAE activations, using event counts for event axes and unique token-carrying participants for patient axes. We summarized categorical discrepancies by total variation distance, $\tfrac12\sum_b|P_f^z(b)-Q_f^z(b)|$, subtracting its expected finite-sample null value to obtain excess TVD. The independent-draw null uses a Kish effective-sample-size approximation for activation-mass summaries; it does not remove within-participant dependence or all confounding (Supplementary Section~\ref{sec:supp_tvd_detail}).

Context summaries compare observed diagnosis-context rates with token-mixture expectations, excluding direct same-token pairs and stratifying temporal context by lag window. Continuous summaries use weighted means and standardized differences with the baseline specified for that field. Patient mass concentration identifies features dominated by a few participants. Detailed descriptor definitions and baseline scopes are in Supplementary Section~\ref{sec:supp_descriptors}; not every metadata annotation uses token adjustment.

Genotype associations were descriptive annotations within the selected variant panel. We correlated participant-level feature activation mass with allele dosage after residualizing both against available sex, birth-timing/age, ancestry-component and batch covariates. Variant QC preceded testing (Supplementary Table~\ref{tab:appendix_genotype_qc}), and Benjamini--Hochberg correction covered the scanned feature--variant family~\cite{benjamini1995controlling}. Selected hits were interpreted alongside token-adjusted dosage differences; they were not a genome-wide or causal analysis (Supplementary Section~\ref{sec:supp_genotypes}).

\subsection{Ruleset interpretation and retraining stability}

Deterministic rules predicted binary feature activation from event identity, date, recording source, demographics, trajectory position and diagnosis history. Training-generated predicates were combined by beam and greedy searches with a complexity penalty; final rule choice used validation participants. We report the scale-invariant mass-weighted Matthews correlation coefficient ($\mathrm{MCC}_{\mathrm{SI}}$). Let $\mu_{f,+}^{\mathrm{train}}$ be feature $f$'s mean positive training activation. Active test positions receive weight $z_{pf}/\mu_{f,+}^{\mathrm{train}}$ and inactive positions weight one. With $T^+$ and $F^-$ the resulting weighted true positives and false negatives, and $FP$ and $TN$ ordinary inactive-position counts,
\begin{equation}
\label{eq:si_mcc}
\mathrm{MCC}_{\mathrm{SI}}=
\frac{T^+TN-FP F^-}{\sqrt{(T^++FP)(T^++F^-)(TN+FP)(TN+F^-)}}.
\end{equation}
We refer to this metric as mass-weighted MCC throughout. The score emphasizes stronger activations while remaining invariant to uniform feature rescaling; it evaluates activation occurrence rather than magnitude. Zero-denominator cases receive zero. Predicate templates, search parameters and evaluation details are in Supplementary Section~\ref{sec:appendix_ruleset_config}.

A separate, locally run language model (\texttt{gpt-oss-120b}) converted aggregate feature cards into natural-language descriptions and confidence categories to assist review, without access to the fitted rules.
We compared these categories with held-out rule recoverability and summary/rule alignment. Language-model confidence was not treated as ground-truth interpretability.

Reproducibility was assessed across three SAE fits with the same frozen transformer and participant assignments. Within-layer matching combined decoder similarity, activation overlap, token and context profiles, and rule agreement. Strong-match connected components accommodated feature splitting or merging; reciprocal best matches supplied pairwise summaries. Recurrence counts distinguish feature instances across runs from matched components, using a support threshold of 100 activations. Matching definitions are in Supplementary Section~\ref{sec:appendix_stability_score}.

\subsection{Input counterfactuals and decoder-direction steering}

Input edits ask what makes a selected feature activate; steering asks how adding or subtracting its decoder direction changes the model’s predictions.
Selected feature interpretations were tested by changing calendar/date information, implied age or diagnosis context, rerunning the unchanged transformer and measuring feature activation at the target position. For L1F68, K29 context was inserted or deleted around K21 events. Deletion applied only to eligible prior context, whereas the card rule describes patient-wide history; both the sampled and actually edited counts are reported.

Steering used 128 test participants with one occurrence of each target diagnosis and 128 matched participants without that code in their available history. The target event, or an other-diagnosis control event, was masked with chapter/block identifiers cleared. Controls were matched on diagnosis token type, age, calendar year and sequence length without selecting on activations or model scores. At the selected feature's native layer output, we applied
\begin{equation}
h'_p=h_p+\alpha q_{75,f}d_f/s(h_p),\qquad -2\leq\alpha\leq2,
\end{equation}
where $d_f$ is its decoder direction and $q_{75,f}$ the 75th percentile of nonzero training activations. The original reconstruction residual and downstream normalization were retained. Signed doses used identical examples; negative doses subtract a reference direction rather than nulling an active feature. We report target-logit changes at $\alpha=\pm1$ in the main figure.

Full dose responses, control comparisons, probability/rank outcomes and paired-bootstrap intervals are in Supplementary Section~\ref{sec:supp_masked_steering}.

\subsection{Retrospective TRD classification and feature interpretation}

The prescription-derived TRD proxy required F32/F33/F34 history, excluded F31, F10--F19 and F20--F29, and counted at least two switches between different adequate antidepressant courses. Courses joined same-agent issue dates separated by at most 14 weeks \cite{fabbri2021genetic}; adequacy required a first-to-last issue span of at least six weeks. Switches required no overlap and $0<s_c-e_p\leq14$ weeks between the preceding end and current start. Participants without an adequate course were excluded. This measures prescribing sequences, not observed non-response or medication adherence; mapping and cohort details are in Supplementary Section~\ref{sec:supp_trd_protocol}.

Models classified the prescription-derived TRD proxy using each participant's complete available diagnosis history, including events that could occur after the qualifying treatment switches. Prescription variables used to define the outcome were excluded from model inputs. We compared models using demographics, diagnosis-code presence, engineered clinical-history features and transformer representations, together with a finetuned bidirectional GRU. Logistic models used L2 regularization, class-balanced weights and training-fitted scaling; validation AUPR selected candidates. Test average precision (reported as AUPR), AUROC and paired model differences used 5,000 participant-bootstrap resamples.

For the primary SAE-based model, each participant was represented by four summaries from each transformer layer: the SAE-reconstructed normalized hidden states at the beginning-of-sequence (BOS), end-of-sequence (EOS) and birth-date (DOB) positions, and their mean across clinical-event positions. We concatenated these summaries and fitted logistic regression. Because SAE decoding is affine, the fitted score can also be written exactly as a weighted sum of the corresponding SAE feature activations, with an adjusted intercept. 
This algebraic re-expression is termed decoder folding: it leaves the fitted score unchanged and does not train a second classifier.
This conversion uses $\gamma=W_D^\top q$, where $q$ is the logistic coefficient vector after accounting for input scaling. We report coefficients per training standard deviation of each pooled SAE feature and their medians and 5th--95th percentiles across 200 bootstrap refits of the training participants. Full details and a separate model fitted directly to SAE features are provided in Supplementary Sections~\ref{sec:supp_trd_models} and~\ref{sec:appendix_trd_decoder_fold}.

Large coefficients do not necessarily  produce large changes in the fitted score when correlated inputs offset one another. We therefore ranked supported SAE inputs by the absolute median standardized coefficient, grouped strongly correlated inputs, and checked their combined contributions in training and validation participants. A displayed input had to retain its coefficient sign across refits, while its group's contribution had to remain reproducible and show limited cancellation internally and against the rest of the score. We displayed the highest-ranked member of each of the first ten qualifying groups. Thresholds and sensitivity checks are given in Supplementary Section~\ref{sec:appendix_trd_decoder_fold}.

\section{Results}

\subsection{Sparse decomposition preserves model representations}

The inspected transformer contained 2.23 million parameters and a 1,133-token vocabulary. At masked diagnosis positions in the test set, top-1 accuracy was 18.3\% and mean reciprocal rank 0.286. These evaluate the masked subset rather than all selected training-objective positions, which also include random replacements and unchanged targets. Architecture and full training comparisons are reported in the Supplement (\ref{sec:supp_mlm_detail}).

\begin{figure}[pos=!htbp]
\centering
\includegraphics[width=\linewidth]{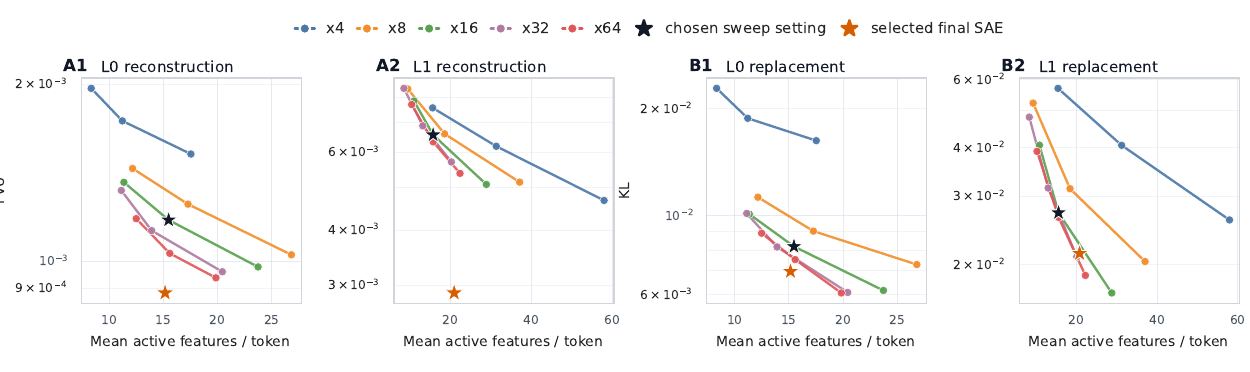}
\caption{\textbf{SAE reconstruction and replacement tradeoff.} Original sweep checkpoints and selected components were evaluated on 100,000 validation tokens, including boundary, DOB and clinical-event positions. Panels show FVU and forward KL divergence against mean active features. Black stars denote sweep choices; orange stars denote the final layer-specific fits.}
\label{fig:sae_pareto_tradeoff}
\end{figure}

The two SAEs yielded 7,712 live features from 8,192 dictionary elements, with 7,167 features meeting the atlas display threshold of at least 10 total activations. On a common 100,000-position validation sample, layers 0 and 1 used averages of 15.1 and 20.9 active features per token. Reconstructions explained more than 99.7\% of hidden-state variance in each layer and preserved 98.6\% and 98.1\% of top model predictions, respectively (Fig.~\ref{fig:sae_pareto_tradeoff}; Table~\ref{tab:model_sae_summary}). This establishes a sparse representation with high reconstruction and replacement fidelity.

\begin{table}[pos=!htbp]\centering\small
\caption{\textbf{Current SAE atlas and reconstruction fidelity.} Evaluation rows use the same 100,000 all-position validation tokens as Fig.~\ref{fig:sae_pareto_tradeoff}; full-export liveness uses a larger observation set. The frozen MLM has two layers, hidden dimension 256, four attention/KV heads, vocabulary 1,133 and 2,234,368 parameters.}
\label{tab:model_sae_summary}
\begin{tabular}{>{\raggedright\arraybackslash}p{\dimexpr0.6\linewidth-2\tabcolsep\relax}>{\raggedleft\arraybackslash}p{\dimexpr0.2\linewidth-2\tabcolsep\relax}>{\raggedleft\arraybackslash}p{\dimexpr0.2\linewidth-2\tabcolsep\relax}}
\toprule
Quantity & Layer 0 & Layer 1 \\
\midrule
Dictionary features & 4,096 & 4,096 \\
Live features, full atlas export & 3,663 & 4,049 \\
Mean active features, evaluation & 15.1 & 20.9 \\
FVU, evaluation & 0.088\% & 0.288\% \\
Replacement top-1 agreement & 98.6\% & 98.1\% \\
Forward KL divergence & 0.00696 & 0.02135 \\
\bottomrule\end{tabular}\end{table}

\subsection{Rulesets and qualitative summaries provide complementary interpretation checks}
\label{sec:results_ruleset_checks}

\begin{figure}[pos=!htbp]
\centering
\includegraphics[width=\linewidth]{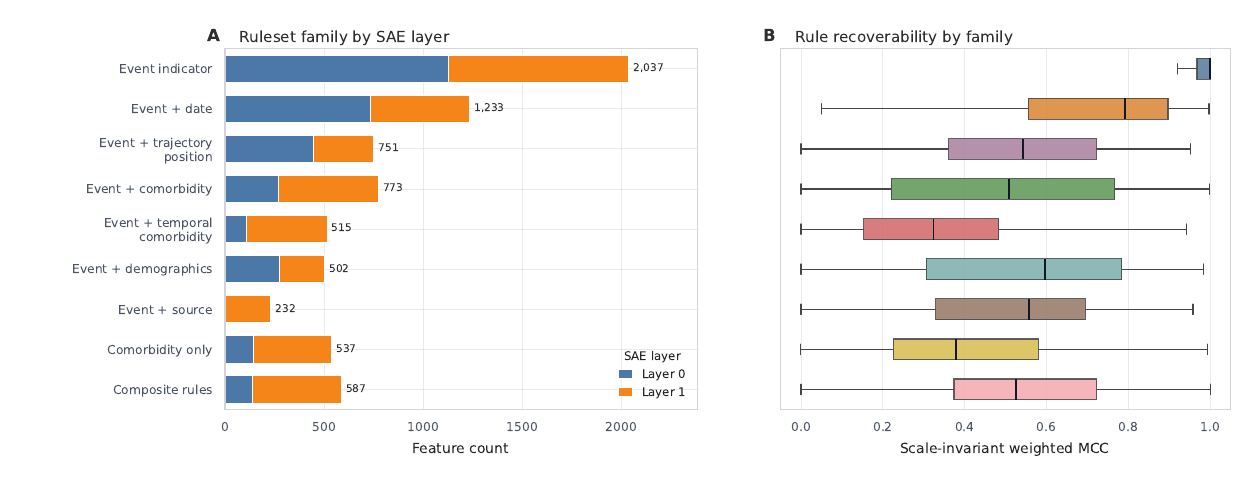}
\caption{\textbf{Rule-family composition and recoverability.} The 7,167 displayed features have at least 10 total activations. (A) Feature counts by rule family and SAE layer; labels give totals across layers. (B) Held-out mass-weighted MCC for the same families, aligned with panel A. Boxes show medians and interquartile ranges; whiskers extend to observations within 1.5 interquartile ranges, with outliers omitted. Active-position weights are activation divided by the feature's mean positive training activation; inactive-position weights are one; the rules predict activation occurrence, not magnitude. Confidence and summary-alignment comparisons are shown in Supplementary Fig.~\ref{fig:supp_ruleset_summary_confidence}.}
\label{fig:ruleset_autointerp}
\end{figure}

Across the 7,167 features meeting the atlas support threshold, compact rules achieved a median held-out mass-weighted MCC of 0.699. Recoverability varied across the eight predicate-defined families and composite rules (Fig.~\ref{fig:ruleset_autointerp}). Event indicators formed the largest family (2,037 features) and were almost perfectly recoverable (median MCC 1.00), followed by event-date rules (0.79). History-dependent families had lower median recoverability: 0.51 for event-comorbidity and 0.32 for event-temporal-comorbidity rules. These results quantify how well the chosen predicate library describes feature activation, rather than establishing clinical correctness.

Rule-family composition also differed by layer. Comorbidity, temporal-comorbidity and comorbidity-only rules together accounted for 16.0\% of supported layer-0 features and 33.3\% of supported layer-1 features (Fig.~\ref{fig:ruleset_autointerp}A). Recording-source rules occurred only in layer 1, and composite rules were also proportionally more common there. These within-layer comparisons account for the different numbers of supported features in the two dictionaries.

Language-model summaries provided a complementary review aid. Median MCC increased from 0.56 for summaries rated ``plausible but uncertain,'' to 0.77 for ``well supported'' and 1.00 for ``highly specific and well supported.'' Higher confidence also coincided with closer summary--ruleset agreement (Supplementary Fig.~\ref{fig:supp_ruleset_summary_confidence}), supporting its use for triage rather than as independent validation. Ruleset recovery itself remains incomplete evidence: activation strength affects the metric's weighting, but the rule predicts activation occurrence rather than magnitude. Even near-perfect recovery therefore leaves open how activation mass is distributed across clinical contexts, motivating the source-token-adjusted analysis below.

\subsection{Feature-card summaries reveal context beyond activation rules}
\label{sec:results_atlas_context}

\begin{figure}[pos=!htbp]
\centering
\includegraphics[width=\linewidth]{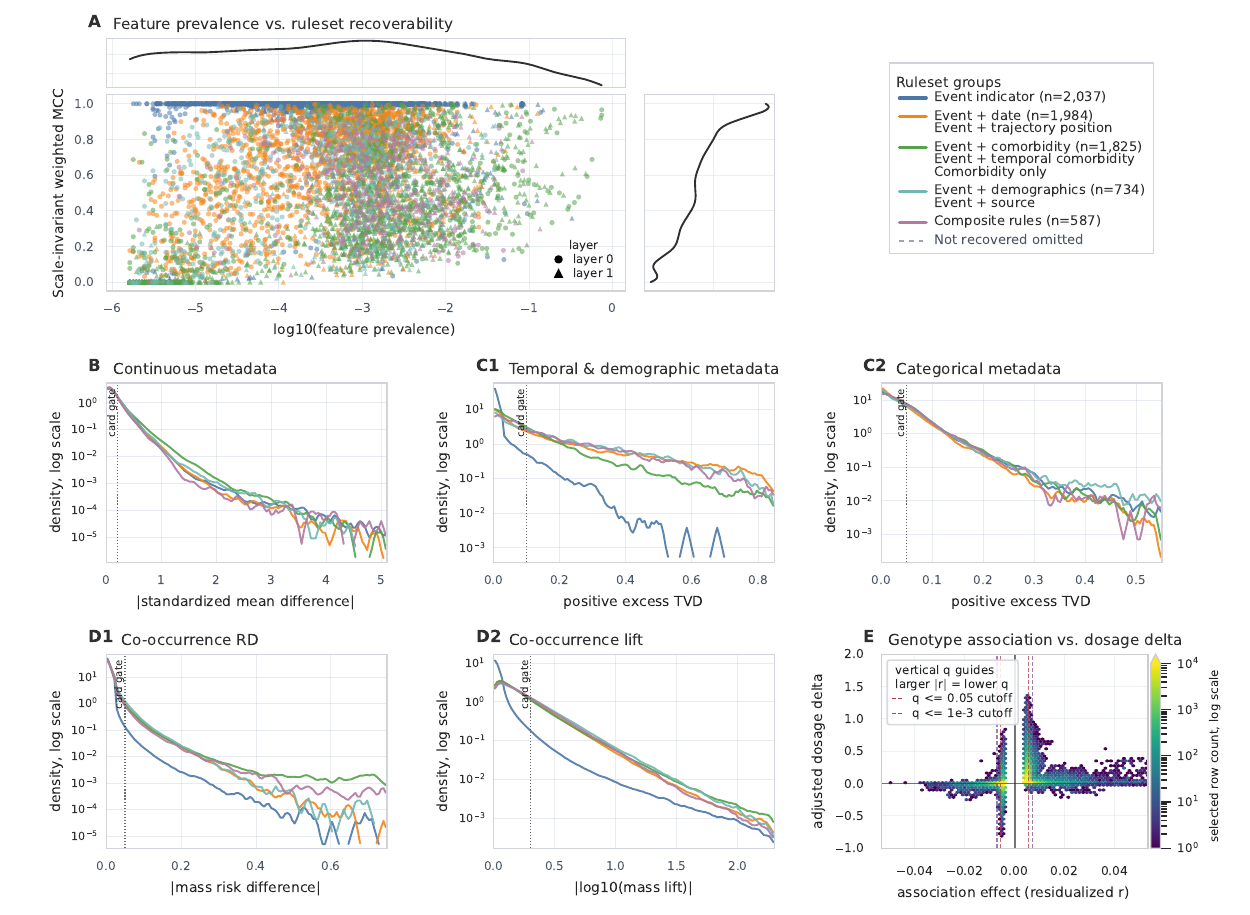}
\caption{\textbf{Atlas-wide feature selectivity exhibits long-tailed distributions.} Colours identify the five rule-family groups shown in the legend. (A) Feature prevalence versus held-out mass-weighted MCC; circles and triangles denote layers 0 and 1, with marginal distributions. (B) Absolute standardized mean differences for continuous metadata. (C1--C2) Positive excess TVD for temporal/demographic and categorical metadata. (D1--D2) Absolute co-occurrence risk differences and log10 lift. (E) Residualized genotype association versus token-adjusted dosage difference, coloured by selected-row count on a log scale; vertical lines are empirical guides for $q\leq0.05$ and $q\leq10^{-3}$. Genotype rows are retained top hits, not a full-scan distribution. Baseline definitions are given in Methods; context and genotype breakdowns appear in Supplementary Figs.~\ref{fig:atlas_context_appendix} and~\ref{fig:atlas_genotype_appendix}.}
\label{fig:feature_atlas_overview}
\end{figure}

Feature-card summaries extended the rule-based description by showing how activation mass was distributed across clinical contexts. Feature prevalence spanned several orders of magnitude (Fig.~\ref{fig:feature_atlas_overview}A). Source-token-adjusted temporal and comorbidity distributions were concentrated near small deviations, with tails of larger differences from expectation (Fig.~\ref{fig:feature_atlas_overview}C1, D1--D2). Here, the expectation combines each activating diagnosis's usual context, weighted by its share of feature activation mass. These contrasts distinguish additional context selectivity from patterns expected from the diagnosis mixture, without removing every source of confounding. Continuous and categorical metadata provide complementary summaries using the field-specific baselines described in Methods.

A diagnosis-indicator rule could describe when a feature activated without fully describing its clinical context. Among 2,037 event-indicator features, 12.2\% exceeded the temporal/demographic excess-TVD threshold of 0.10 and 17.5\% exceeded an absolute co-occurrence risk difference of 0.05 (Supplementary Fig.~\ref{fig:features-above-card-gate}). Thus, even features with simple activation rules could place disproportionate activation mass on particular contexts. Rules provide useful descriptions of activation occurrence, while observed--expected summaries reveal additional selectivity in activation strength.

Genotype annotations complemented the clinical descriptions of features (Fig.~\ref{fig:feature_atlas_overview}E). We compared covariate-adjusted correlations between participant-level feature activation mass and allele dosage with differences between activation-weighted mean dosage and its diagnosis-mixture expectation. Many displayed associations had mean dosages close to this expectation, consistent with enrichment already present among the activating diagnoses. Other pairs showed larger departures, highlighting candidates for investigating genetic associations beyond diagnosis identity. These dosage contrasts are descriptive: they do not quantify the proportion of correlation explained by diagnosis composition or establish an additional association statistically. The displayed pairs comprise the top 50 associations per feature from the selected variant panel.

\subsection{Highlighted features demonstrate interpretation using the atlas}

\begin{figure}[pos=!htbp]
\centering
\includegraphics[width=\linewidth]{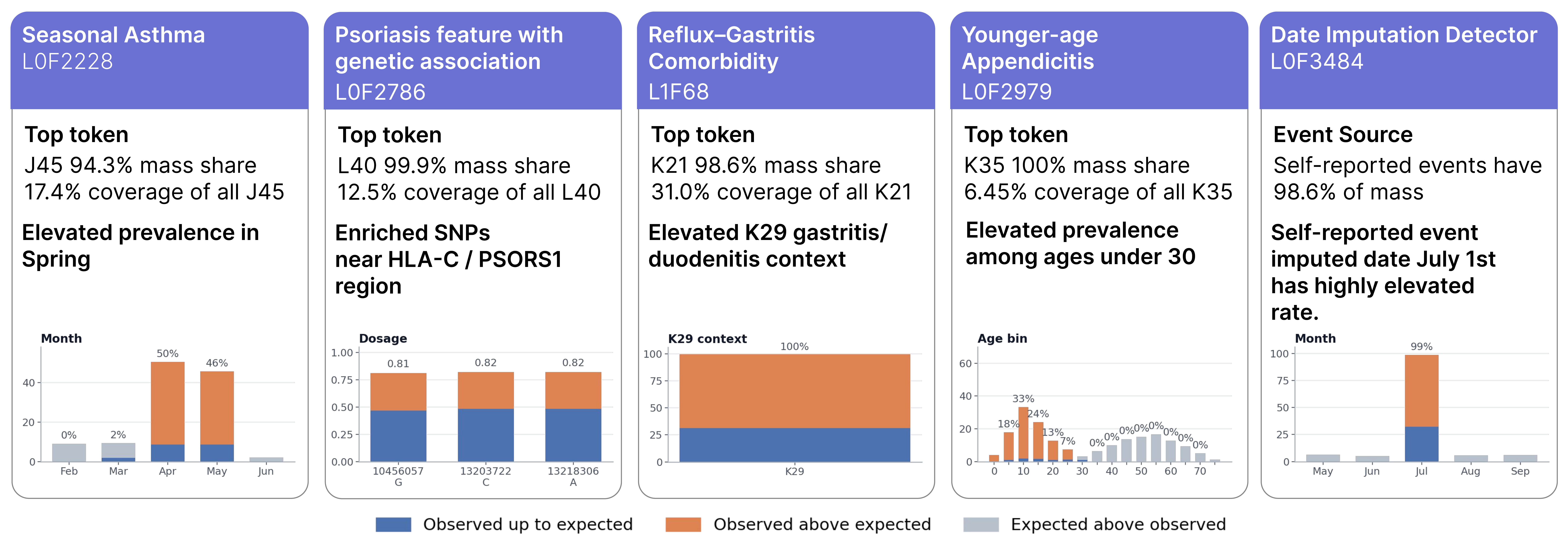}
\caption{\textbf{Selected features illustrate clinical context and recording patterns.} The five cards show seasonal asthma, psoriasis-associated genotype annotations, reflux with gastritis/duodenitis context, younger-age appendicitis and imputed self-report dates. Token mass share is the fraction of feature activation mass carried by a diagnosis; token coverage is the fraction of that diagnosis's events that activate the feature. Observed--expected comparisons use the field-specific baselines described in Methods.}
\label{fig:feature_highlights}
\end{figure}

\begin{figure}[pos=!htbp]
\centering
\includegraphics[width=\linewidth]{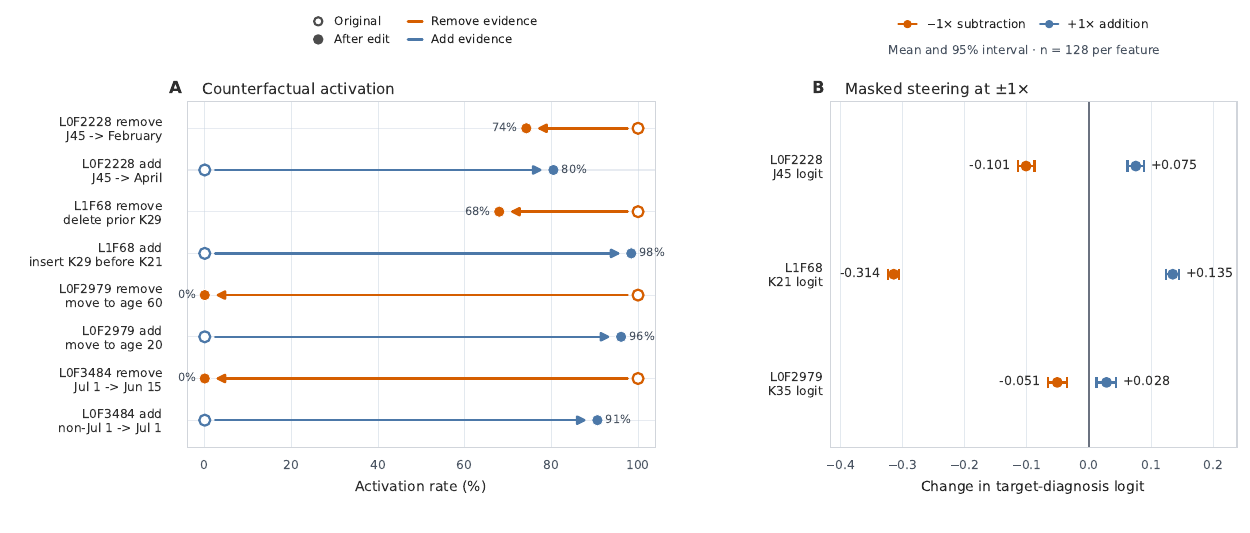}
\caption{\textbf{Input-counterfactual sensitivity and signed masked steering.} (A) Activation rates before (open circles) and after (filled circles) input edits; orange removes evidence and blue adds it. (B) Mean target-logit changes at $-1$ (orange) and $+1$ (blue) reference doses. One reference dose is the 75th percentile of nonzero training activations. Error bars show pointwise 95\% paired participant-bootstrap intervals.}
\label{fig:feature_highlight_interventions}
\end{figure}

Five feature cards illustrate seasonality, genotype enrichment, comorbidity, age and recording artifacts (Fig.~\ref{fig:feature_highlights}). They were selected to demonstrate distinct interpretation modes, and all five had matched counterparts across three independent SAE fits. Targeted input edits tested sensitivity to the proposed context, while decoder-direction steering separately tested effects on masked diagnosis predictions (Fig.~\ref{fig:feature_highlight_interventions}). These were manual checks of selected highlights, outside the automatic annotation pipeline.

L0F2228 captures spring-enriched asthma. J45 events carry 94.3\% of its activation mass, with April and May enriched relative to the source-token baseline. Moving active J45 events to February reduced activation from 100\% to 74.2\%; moving inactive J45 events into April activated the feature in 80.5\% of edited positions. This supports sensitivity to seasonal context within the same diagnosis.

L0F2786 illustrates genotype-linked evidence. L40 psoriasis events carry 99.9\% of its activation mass, and selected genotype associations map near HLA-C/PSORS1, an established psoriasis-susceptibility region~\cite{nair2000localization,fan2008fine}.  The locus annotation is consistent with the psoriasis interpretation; whether this feature distinguishes a genetic subgroup within psoriasis is not established by the displayed association.

L1F68 captures reflux with gastritis/duodenitis context. The question here is whether the model represents reflux differently when gastritis/duodenitis is also recorded elsewhere in the patient’s available history, not whether the two conditions are clinically associated. K21 events carry 98.6\% of its activation mass; K29 context accounts for 99.6\% of mass versus a 31.2\% token-mixture expectation. The feature activates in 19,291 participants (3.84\% of the atlas cohort) and at 31.0\% of K21 events. The rule \texttt{token is K21 AND patient has K29} achieved held-out mass-weighted MCC 0.988. Inserting prior K29 activated the feature in 98.4\% of 128 edited records. Deletion reduced activation from 100\% to 68.0\% across 128 sampled records, but only 42 contained eligible prior K29 and were edited; 86 were unchanged. The card describes patient-wide context, whereas deletion tested prior context only.

L0F2979 captures younger-age appendicitis. Its activation mass is concentrated on K35 events before age 30, with little mass at ages 30--34 and none in older bins, relative to the K35 baseline. The rule \texttt{token is K35 and age <= 30} recovered activation well on held-out data. Moving active K35 events to age 60 eliminated activation in all 51 edited records; moving older inactive events to age 20 activated the feature in 96.1\% of 128 edited records. These edits support age selectivity within appendicitis events.

L0F3484 captures a recording artifact. Self-reported events carry 98.6\% of its activation mass, with strong concentration on July 1, consistent with imputed dates. Moving active July 1 events to June 15 eliminated activation; moving inactive events to July 1 activated the feature in 90.6\% of edited positions. This illustrates how the atlas exposes systematic recording structure alongside clinical context. This matters for clinical interpretation because an apparent calendar pattern can reflect date assignment rather than disease timing.

Adding the selected feature directions increased the model’s scores for asthma, reflux and appendicitis, while subtracting them decreased those scores (Fig.~\ref{fig:feature_highlight_interventions}B). These directions therefore influence predictions in a manner consistent with their clinical descriptions. However, additions also increased the corresponding diagnosis scores at positions containing other diagnoses, indicating that their effects were not restricted to the target context. At the masked positions, these features already had zero activation. Subtracting their directions therefore tested the reverse of adding them, rather than switching off an active feature. The appendix shows how the effects varied with intervention strength, including comparisons with other-diagnosis controls (Supplementary Section~\ref{sec:supp_masked_steering}).

\subsection{Feature stability across SAE retrains}

\begin{figure}[pos=!htbp]
\centering
\includegraphics[width=\linewidth]{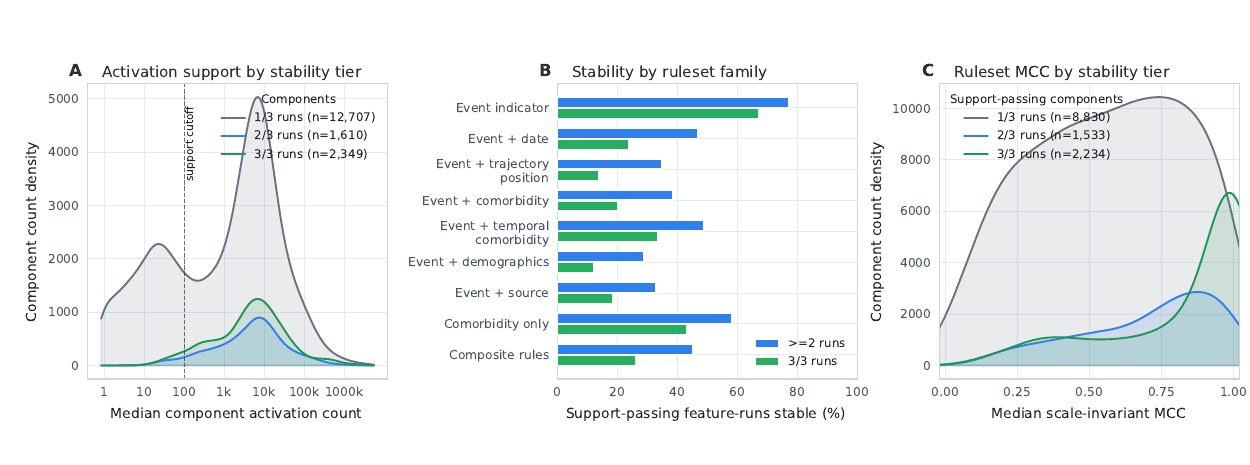}
\caption{\textbf{Feature stability across independent SAE fits.} (A) Component-count-scaled density of median activation count by recurrence tier, including all components; the dashed line marks the support cutoff of 100 activations. (B) Recurrence rates by ruleset family among supported feature instances. (C) Component-count-scaled density of median held-out mass-weighted MCC among support-passing components. Recurrence counts the number of fits represented in a matched component. Panels A and C use different support populations; detailed supported counts appear in Supplementary Table~\ref{tab:supp_stability_recurrence}.}
\label{fig:stability}
\end{figure}

We assessed feature recurrence across three independent SAE fits using the same transformer checkpoint and participant split. Within each layer, features were matched using decoder, activation and feature-card similarity, with strong matches linked into cross-fit components (Appendix~\ref{sec:appendix_stability_score}). Out of the 18,802 feature instances with at least 100 activations, 9,972 (53.0\%) were present in at least two fits and 6,928 (36.8\%) were present in all three runs (Fig.~\ref{fig:stability}). The feature instances formed a total of 12,597 components. The exact component and matching statistics are reported in Supplementary Table~\ref{tab:supp_stability_recurrence} and Appendix~\ref{sec:appendix_stability_score}.

Recurrence varied with activation support and rule family. Activation-count distributions overlapped, while groups observed in only one fit also included substantial low-support density (Fig.~\ref{fig:stability}A). Event indicators were the most recurrent family: 77.2\% of supported fitted features had counterparts in at least one other fit and 67.0\% belonged to groups spanning all three. Comorbidity-only features also recurred relatively often (58.3\% in at least two fits), while other contextual families were less consistently recovered (Fig.~\ref{fig:stability}B). Matched groups spanning more fits generally had higher mass-weighted MCC (Fig.~\ref{fig:stability}C). All five highlighted features in Fig.~\ref{fig:feature_highlights} had matched counterparts in all three fits, including the reflux-with-K29 rule for L1F68.

\subsection{Baseline comparisons support the SAE feature atlas}

\begin{figure}[pos=!htbp]
\centering
\includegraphics[width=\linewidth]{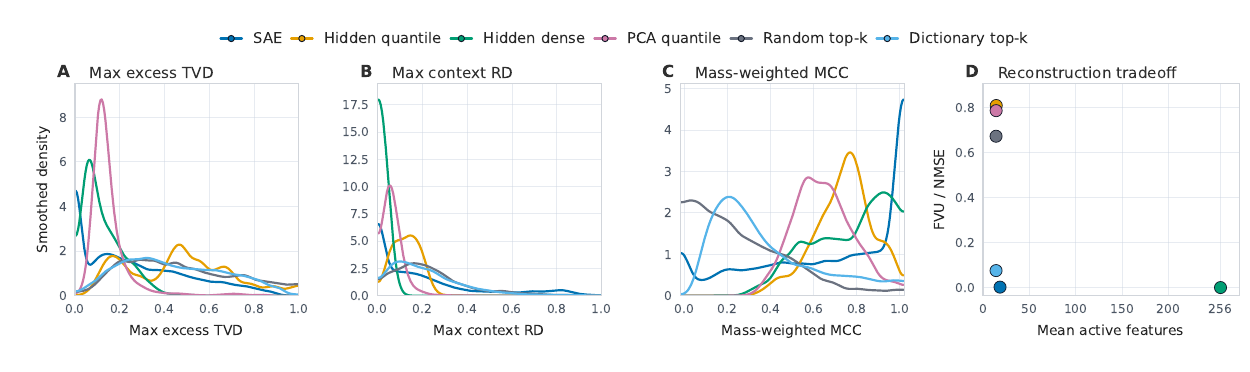}
\caption{\textbf{Representation selectivity, rule recoverability and reconstruction.} (A--C) Smoothed distributions of each feature's maximum excess temporal/demographic TVD, maximum absolute context risk difference and held-out mass-weighted MCC. (D) Reconstruction error (fraction of variance unexplained, FVU) versus mean active features per token-layer. Summary statistics are in Appendix Table~\ref{tab:appendix_baseline_comparison}.}
\label{fig:baseline_interpretability_summary}
\end{figure}

We compared the SAE with a dense hidden-state reference and four sparse alternatives: thresholded hidden-state coordinates, thresholded PCA directions, random top-$k$ directions and learned dictionary top-$k$ features. Sparse representations were evaluated through the same feature-card and ruleset pipeline (Fig.~\ref{fig:baseline_interpretability_summary}). 

The dense reference reconstructs the hidden state exactly; its individual coordinates can be examined, but do not provide sparse activation patterns. Among sparse representations, thresholded hidden coordinates had the highest median mass-weighted MCC (0.751 versus 0.699 for the SAE) and stronger median temporal/demographic selectivity, but high reconstruction error (FVU 0.809). PCA quantiles also reconstructed poorly (FVU 0.786), with lower median recoverability (0.636) and weaker context selectivity. Random top-$k$ features showed broad context enrichment but combined high FVU (0.672) with lower median MCC (0.220). Learned dictionary features reconstructed substantially better (FVU 0.075), although their median MCC (0.300) and reconstruction fidelity remained below the SAE's. The SAE had the lowest sparse-representation FVU (0.002), alongside substantial ruleset recoverability.

Selectivity alone did not identify the most faithful representation: random and dictionary features exceeded the SAE on some median enrichment measures, while thresholded hidden coordinates led median rule recoverability. High reconstruction error limits how faithfully a sparse code represents the hidden state, without establishing that its individual features are uninformative. Among the evaluated configurations, the SAE combined low reconstruction error with sparse, individually inspectable features and substantial ruleset recoverability, supporting its use for the atlas. Full distributions and summary statistics are provided in Appendix~\ref{sec:appendix_representation_baselines}, Table~\ref{tab:appendix_baseline_comparison}.

\subsection{TRD classification illustrates downstream interpretation}
We used treatment-resistant depression (TRD) classification to test whether a downstream score could be traced to atlas features, and compare the performance of the SAE feature based classifiers with traditional baselines. The prescription-derived proxy identified 968 positives among 14,905 eligible participants (6.5\%). The analysis used full available diagnosis trajectories and therefore measures retrospective phenotype discrimination, rather than prospective treatment response. Cohort composition and label construction are reported in Supplementary Section~\ref{sec:supp_trd_protocol}. The SAE-feature based classifier uses the folded decoder as described in Methods. Since a feature can be active for multiple tokens of the input sequence the classifier input differentiates the source. For example, the classifier assigning a coefficient to L1F3442 EOS, means this feature is relevant to classification when activated on EOS tokens.

\begin{figure}[pos=!htbp]
\centering
\includegraphics[width=\linewidth]{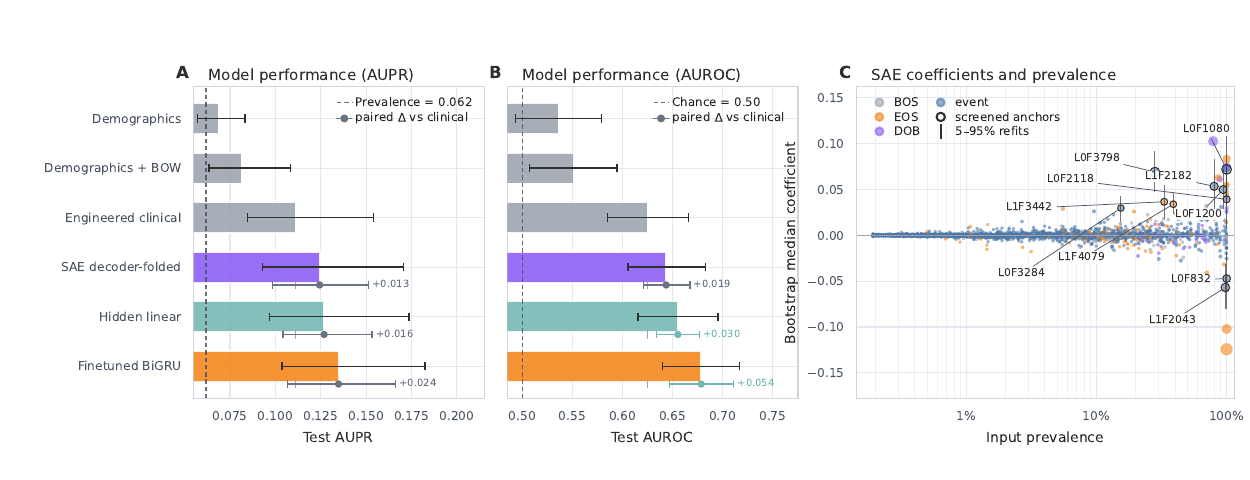}
\caption{\textbf{TRD discrimination and SAE coefficient prevalence.} (A--B) Test AUPR and AUROC with 95\% paired participant-bootstrap intervals. Smaller points beneath the learned-representation bars show paired differences from engineered clinical history, translated onto the performance axis; the short tick marks zero difference. Teal intervals exclude zero and grey intervals include it. Dashed lines indicate test prevalence and chance AUROC. (C) Training prevalence versus bootstrap-median standardized decoder-folded coefficient. Colours identify BOS, EOS, DOB and mean clinical-event inputs; point size encodes coefficient magnitude and opacity is uniform. Open circles and feature identifiers mark the ten displayed inputs, with 5th--95th refit intervals.}
\label{fig:trd_summary}
\end{figure}

On the held-out test set, the SAE classifier achieved AUPR 0.124 and AUROC 0.643, compared with 0.111 and 0.625 for engineered clinical-history features, comprising diagnosis presence and demographics plus summaries of diagnosis burden, event timing and recording source. Hidden-state and finetuned BiGRU comparators had AUPR point estimates of 0.127 and 0.135 (Fig.~\ref{fig:trd_summary}A--B). The SAE classifier had only slightly lower point estimates than the BiGRU and hidden-state logistic models, while still having a higher point estimate than the engineered clinical model. Bootstrapped confidence intervals overlap and certain superiority cannot be established. Exact estimates and paired differences appear in Supplementary Tables~\ref{tab:trd_corrected_models} and~\ref{tab:supp_trd_paired_differences}.

For interpretation, we ranked inputs by absolute standardized coefficient and applied dependence-aware screening, selecting ten inputs for display (Fig.~\ref{fig:trd_summary}C; Table~\ref{tab:trd_screened_sae_groups}). Their feature descriptions include anxiety, lower respiratory infection history, back-pain history and obesity, alongside primary-care recording, event dates and broader diagnosis mixtures. Seven of the ten features recurred in at least one additional SAE fit. The selection was largely unchanged across screening variants (Supplementary Table~\ref{tab:supp_screen_sensitivity}); the screening procedure and contribution diagnostics are detailed in Appendix~\ref{sec:appendix_trd_decoder_fold}.

\begin{table}[pos=!htbp]
\centering
\caption{\textbf{Selected SAE inputs and suggested interpretations.} Coefficients are bootstrap medians per training standard deviation. SAE recurrence counts fits containing a matched counterpart of the underlying feature out of three; it does not measure TRD-coefficient stability. Event denotes mean clinical-event activation, EOS the sequence-end position and DOB the birth-date position. Selection accounts for correlated inputs as described in Methods and Appendix~\ref{sec:appendix_trd_decoder_fold}.}
\label{tab:trd_screened_sae_groups}
\small
\begin{tabular}{>{\raggedright\arraybackslash}p{\dimexpr0.24\linewidth-2\tabcolsep\relax}>{\raggedleft\arraybackslash}p{\dimexpr0.13\linewidth-2\tabcolsep\relax}>{\centering\arraybackslash}p{\dimexpr0.17\linewidth-2\tabcolsep\relax}>{\raggedright\arraybackslash}p{\dimexpr0.46\linewidth-2\tabcolsep\relax}}
\toprule
Input & Coeff. & SAE recurrence & Suggested interpretation \\
\midrule
L0F1080-event & $+0.072$ & 3/3 & Skin/eye-infection-weighted diagnosis mix \\
L0F3798-event & $+0.069$ & 3/3 & Anxiety-event signal (F41) \\
L1F2043-event & $-0.057$ & 1/3 & Primary-care-enriched diagnosis recording \\
L1F2182-event & $+0.054$ & 3/3 & Lower respiratory infection history (J22) \\
L0F1200-event & $+0.050$ & 3/3 & 1997--2005 event-date component \\
L0F832-event & $-0.047$ & 1/3 & Skin/airway-weighted diagnosis mix \\
L0F2118-DOB & $+0.039$ & 3/3 & Pain/multimorbidity context at DOB \\
L1F3442-EOS & $+0.037$ & 2/3 & Back-pain history (M54) \\
L1F4079-EOS & $+0.034$ & 1/3 & Anxiety-enriched psychiatric/cognitive context \\
L0F3284-event & $+0.030$ & 3/3 & Obesity diagnosis (E66) \\
\bottomrule
\end{tabular}
\end{table}

\FloatBarrier
\section{Discussion}

TRAIL-UKB demonstrates a method for examining how a clinical trajectory transformer represents diagnoses together with their surrounding context. Across 7,712 live SAE features, the atlas exposed diagnosis indicators, contextual patterns and recording-related structure. The methodological contribution is to characterize these features against explicit diagnosis-conditioned expectations and connect their descriptions to complementary tests of representation fidelity, activation behaviour, recurrence and model output. This extends the application of sparse dictionaries to medical interpretation~\cite{wu2024beyond} by addressing a problem specific to clinically meaningful event tokens: an apparently informative feature description may simply restate the usual properties of the diagnoses on which it activates. The contribution lies in the formulation and evaluation of this characterization procedure using established decomposition and analysis tools.

The feature atlas contains a long tail of features, whose selectivity cannot be attributed to their activating diagnoses. One highlighted feature is an imputed date encoding feature that activates on events occurring on July 1st (the date of self-reported diagnoses). Another example feature marks reflux (K21) events with gastritis/duodenitis (K29) context. The functionality of the highlighted feature is validated through input counterfactuals and masked steering. Changing event dates, implied age or diagnosis context validates that the proposed evidence affects feature activation. Injecting a decoder direction shows the feature's effect on the remaining model computation and output. 

A useful feature atlas must preserve the model representation while offering descriptions that can be examined and reproduced. The combination of high reconstruction fidelity, recoverable activation rules and recurrence across SAE fits supports using TRAIL-UKB for this purpose. The baseline comparison shows why these criteria should be considered together: features can have simple rules or strong contextual enrichment while reconstructing the hidden state poorly. The SAE offered a useful combination at the evaluated sparsities, although substantial variation across retrains means that its dictionary should not be treated as a unique decomposition.

The TRD example shows how atlas features can make a classifier's associations easier to examine. Its modest discrimination did not establish an improvement over engineered clinical-history features, but its score could be expressed exactly through features describing anxiety, respiratory infection, back pain and obesity alongside calendar and recording-source patterns. This allows clinicians and model developers to examine whether the associations are clinically plausible or reflect how histories were recorded.

The study concerns one compact transformer and one cohort, using first-occurrence diagnoses that omit repeated events and much of the severity and treatment context available in routine care. The TRD analysis used a prescription-derived proxy and complete available histories, including potentially post-switch diagnoses; it therefore does not establish prospective prediction of treatment response. Genotype annotations came from a selected variant panel, and automated descriptions require clinical scrutiny. External validation is needed to assess whether feature meanings persist across datasets, while a clinician-reader study would be needed to test whether the cards improve interpretation or model review compared with simpler summaries.

TRAIL-UKB provides candidate features for subsequent circuit analysis of how clinical and recording-related information is combined across layers to influence predictions~\cite{marks2024sparse,lindsey2025biology}. Connecting these features into tested computational pathways could support deeper mechanistic inspection and, with clinical assessment and external validation, more systematic auditing of medical models.

\subsection{Statistical reporting and software availability}

Where formal multiple-testing correction was required, we used Benjamini--Hochberg false-discovery rates within predefined analysis families. Support and privacy gates were applied before feature-card rendering and before using aggregate rows as evidence for claims about manuscript interpretation. Unsupported rows could be retained as internal diagnostics but were not used as evidence in the manuscript.

The analysis pipeline covers sequence export, tokenizer training, Delphi-MLM training and evaluation, hidden-state export, SAE fitting, replacement evaluation, feature-descriptor generation, ruleset benchmarking, genotype summaries, representation-baseline comparisons, and downstream classifiers. Code, command templates, schemas, and privacy-screened aggregate artifacts will be released in the project repository where permitted by UK Biobank policy. This includes rendered feature cards for features with more than 100 activating positions, after applying row-level support and privacy gates. Raw UK Biobank exports, participant-level timelines or metadata, participant identifiers, participant-linked sparse activation rows, protected paths, and credentials will not be redistributed.

\section*{Data availability}

Individual-level UK Biobank data cannot be redistributed by the
authors. Researchers may apply for access through the UK Biobank
research-access process. Aggregate outputs described in this work
will be made available where permitted.

\section*{Code availability}

Code implementing the analysis pipeline will be made publicly
available upon publication.

\subsection{Ethics statement}
This work uses data provided by patients and collected by the NHS as part of their care and support. Copyright © (2019), NHS England. Reused with the permission of the UK Biobank (Application Number 1602). All rights reserved. Individual-level records were accessed only within the governed research environment and were not redistributed. The released materials contain only source code, command templates, schemas, model-analysis outputs, and privacy-screened aggregate artifacts.

\section{Competing interests}
This work was supported by the National Research, Development and Innovation Office of Hungary (NKFIH) under grant OTKA-K139330. GJ was supported by the the Hungarian Brain Research Program (Grant: 2017-1.2.1-NKP-2017-00002; NAP2022-I-4/2022), by the National Research, Development and Innovation Office of Hungary (NKFIH K 143391), and by the Thematic Excellence Program. Project TKP2021-EGA-25 (Ministry of Innovation and Technology of Hungary).

\appendix
\section{Data}

\subsection{Field inventory}
\label{sec:supp_inventory}
\begin{table}[pos=!htbp]
\caption{\textbf{UK Biobank field groups exported for the analysis.} Counts refer to selected UK Biobank field identifiers in the repository field-list configuration, not to non-missing participant values.}
\label{tab:ukb_field_groups}
\centering
\small
\begin{tabular}{lr}
\hline
Field group & Number of selected fields \\
\hline
Demographics & 3 \\
Anthropometry & 9 \\
Vitals and physical function & 5 \\
Lifestyle questionnaire & 10 \\
Mental-health questionnaire & 11 \\
General health and reproductive health & 6 \\
Blood count & 31 \\
Urine assays & 4 \\
Blood biochemistry & 30 \\
NMR metabolomics & 249 \\
ICD-10 disease onset & 1,127 \\
ICD-10 disease source & 1,127 \\
Death register & 1 \\
\hline
Total selected field IDs & 2,613 \\
\hline
\end{tabular}
\end{table}

\subsection{Event export and preprocessing}
\label{sec:supp_preprocessing}

Raw UK Biobank tabular exports were converted into two participant-level analysis tables: an event table and a patient-metadata table. Participants with missing year or month of birth (\texttt{Field 34}, \texttt{Field 52}) were excluded before event construction because birth date was required for event ages. The event table contains birth, disease first-occurrence and death events and includes absolute dates, age at event and event source for later analysis. For birth dates, the day was imputed to the first day of the recorded birth month. Death dates were derived from age at death (Field \texttt{40007}) and the imputed birth dates. Diseases were sourced from the first-occurrence fields (\texttt{Category 1712}) and use three-character ICD-10 codes, with block and chapter hierarchy joined based on the UK Biobank helper files. Events before birth or after death were clamped, or dropped if outside a 90-day window. For non-disease metadata fields, the export used the first UK Biobank instance and array position only, because of high missingness in others.

\clearpage
\section{MLM training detail}
\subsection{Delphi-MLM input, architecture, and training: technical detail}
\label{sec:supp_mlm_detail}

We represented each event as a discrete token. The vocabulary consisted of special tokens used for training and sequence boundaries, birth and death tokens, and the 1,127 selected three-character ICD-10 first-occurrence codes. We also considered optional context tokens following the Delphi-2M tokenization scheme for sex, BMI, alcohol use, and tobacco use. The final atlas uses the no-context configuration. Full-context candidates achieved somewhat better prediction metrics in the candidate comparison below; the atlas checkpoint therefore should not be interpreted as the prediction-optimal configuration. Here, ``no context'' means omission of these optional context tokens, not removal of calendar or diagnosis-hierarchy information.

Our transformer model was based on nanochat, a minimal implementation of a modern generative language model, but was modified to use bidirectional attention for the BERT-style masked language modelling objective. We further extended the model with additional embedding layers to incorporate temporal and disease-hierarchy information.

For participant $i$ and sequence position $j$, candidate input embeddings combined token, token-type, ICD-10 block, ICD-10 chapter, age and date components. Each component was ablated during model selection. The final checkpoint retained token/type, block/chapter and year/month/day embeddings but disabled the age embedding, as recorded in the supplementary model-ablation table. Its input was
\begin{equation}
\begin{array}{lcl}
e^{\mathrm{tok}}_{ij} &=&
E_{\mathrm{tok}}(x_{ij})+E_{\mathrm{type}}(t_{ij}), \\
e^{\mathrm{hier}}_{ij} &=&
E_{\mathrm{block}}(b_{ij})+E_{\mathrm{chapter}}(c_{ij}), \\
e^{\mathrm{time}}_{ij} &=&
E_{\mathrm{year}}(y_{ij})+E_{\mathrm{month}}(m_{ij})+E_{\mathrm{day}}(d_{ij}), \\
z_{ij} &=& e^{\mathrm{tok}}_{ij}+e^{\mathrm{hier}}_{ij}+e^{\mathrm{time}}_{ij}.
\end{array}
\end{equation}

The masked language modelling objective followed the original BERT-style corruption scheme, but sampling used the tokenizer's event-position flag, which includes date-of-birth events as well as diagnoses. For each input sequence, we sampled 15\% of eligible event positions. Of these sampled positions, 80\% were replaced with a special mask token, 10\% were replaced with a randomly selected token from the event vocabulary, and 10\% were left unchanged. Temporal embeddings were retained at all sampled positions. To prevent leakage of the target ICD-10 code through hierarchy embeddings, ICD-10 chapter and block embeddings were always derived from the corrupted input token rather than from the original target token. Thus, positions replaced by the mask token received the special no-chapter and no-block identifiers used for special and non-disease tokens; positions replaced by birth, death, or another non-disease token received the no-chapter and no-block identifiers; and unchanged positions retained the original chapter and block identifiers. The final cross-entropy loss was computed only over the sampled event positions, with equal weight for each sampled event and no reweighting by disease prevalence.

\begin{equation}
\mathcal{L}_{\mathrm{MLM}}
=
-\sum_i \sum_{j \in M_i}
\log p_\theta(x_{ij}\mid \tilde{\mathbf{x}}_i),
\end{equation}
where $M_i$ denotes the sampled event positions for participant $i$, and $\tilde{\mathbf{x}}_i$ denotes the corrupted input sequence.

We trained the model using AdamW with a learning-rate schedule consisting of a warmup phase followed by cosine decay. We used a weight decay of 0.01, gradient clipping at 1.0, and a batch size of 512 participants. Models were ranked by evaluation loss, and we additionally report top-$k$ accuracy and mean reciprocal rank (MRR). All models were trained with mixed-precision bf16 on an NVIDIA RTX PRO 6000 Blackwell GPU.

\clearpage
\subsection{MLM transformer ablations}

\begin{table}[pos=!h]
\centering
\caption{\textbf{Base configurations for MLM ablation sweeps.} The ablation program used two closely related exploratory configurations. Short sweeps used an 8-layer, 768-hidden grouped-query-attention model with no dropout; the later longer sweeps incorporated the dropout and attention settings shown below. Individual rows in Table~\ref{tab:mlm-ablation-sweep} change the indicated setting.}
\label{tab:mlm-ablation-base-config}
\small
\begin{tabular}{p{0.25\linewidth}p{0.48\linewidth}p{0.20\linewidth}}
\hline
Sweep rows & Fixed settings & Changed setting(s) \\
\hline
5k size sweep & full context; date/age/ICD on; dropout 0; LR 3e-4 & layers / hidden dimension \\
Context, temporal, hierarchy & 8L / 768; 12 / 4 heads; dropout 0; LR 3e-4 & context or embedding fields \\
Learning rate & 8L / 768; 12 / 4 heads; full context; dropout 0 & LR \\
Dropout & 8L / 768; 12 / 4 heads; full context; LR 3e-4 & dropout 0.2 \\
Attention & 8L / 768; full context; dropout 0.2; LR 3e-4 & 12 / 12 MHA \\
50-epoch size sweep & full context; date/age/ICD on; dropout 0.2; LR 3e-4 & layers / hidden dimension \\
\hline
\end{tabular}
\end{table}

\begin{table}[pos=!h]
\centering
\caption{\textbf{MLM ablation sweep.} Validation loss is the minimum validation loss logged in local W\&B history for each run. Bold values mark the lowest validation loss within multi-row sweep blocks.}
\label{tab:mlm-ablation-sweep}
\begin{tabular}{llr}
\hline
Sweep & Changed setting & Best val. loss \\
\hline
5k-step size sweep & 2L / 128 & 3.765 \\
5k-step size sweep & 4L / 256 & 3.587 \\
5k-step size sweep & 8L / 512 & 3.483 \\
5k-step size sweep & 8L / 768 & \textbf{3.441} \\
5k-step size sweep & 12L / 768 & 3.446 \\
\hline
Context & full & \textbf{3.345} \\
Context & no context & 3.352 \\
Context & sex only & 3.348 \\
\hline
Temporal & date+age off & 3.786 \\
Temporal & age only & 3.510 \\
Temporal & date only & \textbf{3.341} \\
\hline
ICD hierarchy & off & 3.338 \\
\hline
Learning rate & 5e-5 & 3.525 \\
Learning rate & 1e-4 & 3.458 \\
Learning rate & 3e-4 & \textbf{3.451} \\
Learning rate & 1e-3 & 3.608 \\
\hline
Dropout & 0.2 & 3.284 \\
\hline
Attention & MHA, 12/12 heads & 3.257 \\
\hline
50-epoch size sweep & 2L / 128 & 3.584 \\
50-epoch size sweep & 4L / 256 & 3.362 \\
50-epoch size sweep & 8L / 512 & 3.291 \\
50-epoch size sweep & 8L / 768 & \textbf{3.272} \\
50-epoch size sweep & 12L / 768 & 3.280 \\
\hline
\end{tabular}
\end{table}

\begin{table}[t]
\centering
\caption{\textbf{Final Delphi-MLM configuration used for the SAE atlas.} The no-context v2 checkpoint (\texttt{final\_run\_2L\_no\_context\_med\_v2}) was used for hidden-state export, SAE training, feature-card construction and downstream analyses. Its separate masked-diagnosis test evaluation yielded top-1 accuracy 0.182687 and MRR 0.286398. The candidate tables below report archived comparison evaluations, including the earlier no-context run, and are not the source of these final-v2 results.}
\label{tab:mlm-final-config}
\begin{tabular}{ll}
\hline
Field & Value \\
\hline
Dataset/context & no context \\
Layers / hidden dimension & 2 / 256 \\
Attention heads / KV heads & 4 / 4 \\
Maximum sequence length & 1,024 \\
Token vocabulary size & 1,133 \\
Parameter count & 2,234,368 \\
Date / age / ICD embeddings & yes / no / yes \\
Dropout & 0.1 \\
Learning rate / minimum learning rate & 3e-4 / 1e-5 \\
Epochs & 100 \\
MLM corruption & 15\%; 80/10/10 mask/random/original \\
\hline
\end{tabular}
\end{table}

\begin{table}[t]
\centering
\small
\caption{\textbf{Final MLM candidate losses under matched evaluation scopes.}
Validation and test losses are computed over all selected MLM prediction positions. Disease losses exclude date-of-birth events. Disease mask, changed, and unchanged columns split selected disease positions by the MLM corruption type.}
\label{tab:mlm-final-selected-losses}
\begin{tabular}{l|r|r|rrrr}
\hline

Variant & Val & Test & Disease test & masked & changed & unchanged \\
\hline
2L/128, full context & 3.525 & 3.528 & 3.913 & 4.253 & 4.726 & 0.363 \\
2L/128, no context & 3.575 & 3.577 & 3.967 & 4.318 & 4.785 & 0.363 \\
2L/256, full context & \textbf{3.435} & \textbf{3.436} & \textbf{3.810} & \textbf{4.154} & \textbf{4.528} & \textbf{0.320} \\
2L/256, no context & 3.486 & 3.492 & 3.873 & 4.226 & 4.614 & 0.323 \\
2L/256, sex context & 3.476 & 3.484 & 3.857 & 4.212 & 4.567 & 0.322 \\
8L/768, full context & 3.440 & 3.444 & 3.819 & 4.170 & 4.539 & 0.345 \\
\hline
\end{tabular}
\end{table}

\begin{table}[]
\centering
\small

\caption{\textbf{Final MLM candidate top-1 accuracy.}}
\begin{tabular}{l|r|r|rrrr}
\hline
Variant & Val & Test & Disease test & masked & changed & unchanged \\
\hline
2L/128, full context & 0.319 & 0.320 & 0.246 & 0.171 & 0.130 & 0.965 \\
2L/128, no context & 0.316 & 0.314 & 0.239 & 0.164 & 0.121 & 0.964 \\
2L/256, full context & \textbf{0.330} & \textbf{0.329} & \textbf{0.256} & \textbf{0.182} & 0.148 & 0.966 \\
2L/256, no context & 0.325 & 0.324 & 0.250 & 0.175 & 0.136 & \textbf{0.967} \\
2L/256, sex context & 0.326 & 0.324 & 0.251 & 0.176 & 0.138 & 0.966 \\
8L/768, full context & 0.329 & 0.328 & 0.255 & 0.181 & \textbf{0.149} & 0.963 \\
\hline
\end{tabular}

\label{tab:mlm-final-selected-top1}
\end{table}

\begin{table}[]
\centering
\small

\caption{\textbf{Final MLM candidate MRR.}}
\begin{tabular}{l|r|r|rrrr}
\hline
Variant & Val & Test & Disease test & masked & changed & unchanged \\
\hline
2L/128, full context & 0.403 & 0.403 & 0.338 & 0.273 & 0.230 & 0.972 \\
2L/128, no context & 0.397 & 0.397 & 0.331 & 0.265 & 0.221 & 0.971 \\
2L/256, full context & \textbf{0.416} & \textbf{0.416} & \textbf{0.352} & \textbf{0.287} & 0.255 & 0.973 \\
2L/256, no context & 0.410 & 0.409 & 0.345 & 0.279 & 0.243 & \textbf{0.974} \\
2L/256, sex context & 0.411 & 0.410 & 0.346 & 0.280 & 0.247 & 0.973 \\
8L/768, full context & 0.415 & 0.415 & 0.351 & 0.286 & \textbf{0.258} & 0.970 \\
\hline
\end{tabular}

\label{tab:mlm-final-selected-mrr}
\end{table}

\clearpage

\section{Sparse autoencoder details}
\subsection{Sparse autoencoder training, architecture, and evaluation: technical detail}
\label{sec:supp_sae_detail}

We exported hidden states of Delphi-MLM from clean, uncorrupted input sequences. For each transformer layer, hooks captured the post-block residual output at all non-padding positions in the train, validation, and test splits. This produced one dense vector for each token--layer pair. Using training-split states only, we then trained one JumpReLU sparse autoencoder (SAE) separately for each transformer layer; validation states were used for checkpoint evaluation, and test states were reserved for held-out analyses.

We use the same column-vector convention as in the main text: $D$ is the hidden dimension, $F$ the dictionary size, $W_E\in\mathbb{R}^{F\times D}$ and $W_D\in\mathbb{R}^{D\times F}$. Each hidden state is normalized before encoding; JumpReLU selects active features, and decoding reconstructs the normalized state:
\begin{equation}
\begin{array}{lcl}
    u &=& W_E(\widetilde h-b_D)+b_E, \\
    z_f &=& u_f\mathbf{1}[u_f>\theta_f],\qquad \theta_f=\exp(\tau_f), \\
    \widehat{\widetilde h} &=& W_Dz+b_D,\qquad \widehat h=\widehat{\widetilde h}/s(h).
\end{array}
\end{equation}
Here $\widetilde h=s(h)h$, with $s(h)=\sqrt{D}/\max(\|h\|_2,10^{-6})$. Decoding is affine in SAE activations in normalized coordinates; restoring raw hidden states additionally requires each input's original scale. In the losses below, $d_f$ denotes column $f$ of $W_D$.

The SAEs balance reconstruction fidelity and the number of active features using a multi-term loss that combines reconstruction error, a decoder-norm-weighted sparsity penalty, and a dead-feature auxiliary loss, which encourages inactive features to cross their JumpReLU thresholds.

\begin{equation}
\begin{array}{lcl}
\mathcal{L}_{\mathrm{dead}}
&=&
\sum_{f=1}^{F}
m_f \, \|d_f\|_2 \,
[\theta_f - u_{f}]_+ , \\
\mathcal{L}_{\mathrm{reconstruction}}
&=&
\|\hat{h} - h\|_2^2 \\
\mathcal{L}_{\mathrm{sparsity}}
&=&
\frac{D}{F}
\sum_{f=1}^{F}
\tanh\!\left(\alpha\, z_{f}\|d_f\|_2\right), \\
\mathcal{L}_{\mathrm{SAE}}
&=&
\mathcal{L}_{\mathrm{reconstruction}}
+
\lambda_s(t) \mathcal{L}_{\mathrm{sparsity}}
+
\lambda_d \mathcal{L}_{\mathrm{dead}} ,
\end{array}
\end{equation}

Here $m_f$ indicates whether feature $f$ was considered dead, meaning that it had not fired within a recent training window, and $[\cdot]_+$ denotes the positive part. The sparsity coefficient $\lambda_s(t)$ was time-dependent: it was ramped from zero to a target value during training, allowing the SAE to first learn a useful reconstruction before receiving the full sparsity pressure.

We evaluated SAEs along three axes: reconstruction fidelity, dictionary sparsity, and model-computation preservation. The final atlas retained the original layer-0 fit (sparsity coefficient 12, JumpReLU backward bandwidth 1.0) and replaced layer 1 with a coefficient-64 fit using bandwidth 0.10. Each layer had 4,096 features and a training budget of 200 million token presentations; batch sizes were 2,048 for layer 0 and 8,192 for the layer-1 retrain. Bandwidth controls the surrogate gradient during training, not the thresholding rule at inference. All layer-1-dependent atlas summaries below use this retrain. Reconstruction fidelity was measured by fraction of variance unexplained (FVU), computed over the set of evaluated positions $\mathcal{P}$:
\begin{equation}
\label{eq:fvu}
\mathrm{FVU} =
\frac{
\sum_{p \in \mathcal{P}}\|h_p-\hat{h}_p\|_2^2
}{
\sum_{p \in \mathcal{P}}\|h_p-\bar{h}\|_2^2
}.
\end{equation}

Dictionary sparsity was summarized by $L_0$, the number of active SAE features per token, $\|z_p\|_0$. We also reported the number of dead features, defined as features with no activations over the evaluation set, and the active-feature fraction, defined as the proportion of dictionary elements active at least once.

Finally, we tested whether SAE reconstructions preserved model computation by rerunning Delphi-MLM after replacing the selected layer outputs with SAE reconstructions. Let $P_p$ denote the original MLM predictive distribution at position $p$, and let $Q_p$ denote the corresponding distribution after SAE replacement. We used forward KL divergence,

\begin{equation}
D_{\mathrm{KL}}(p \| q)
=
\frac{1}{|\mathcal{P}|}
\sum_{p \in \mathcal{P}}
\sum_{v \in \mathcal{V}}
P_p(v)
\log
\frac{P_p(v)}{Q_p(v)}
\end{equation}
as the primary replacement-fidelity metric. This treats the original model as the reference distribution and measures how much the SAE replacement changes its predictions. We also reported top-1 agreement, the fraction of positions for which the most probable token was unchanged:
\begin{equation}
\mathrm{Top1Agreement}
=
\frac{1}{|\mathcal{P}|}
\sum_{p \in \mathcal{P}}
\mathbf{1}
\left[
\arg\max_{v} P_p(v)
=
\arg\max_{v} Q_p(v)
\right].
\end{equation}

\subsection{SAE training configuration}

\begin{table}[pos=!htbp]
\centering
\caption{\textbf{Final SAE training configuration.} Hyperparameters for the sparse autoencoders used in the final feature atlas. Sparse autoencoders were trained separately for each selected MLM layer using the stored final MLM checkpoint.}
\label{tab:sae-training-config}
\begin{tabular}{ll}
\hline
Setting & Value \\
\hline
SAE layers & Transformer layers 0 and 1 \\
Expansion factor & $16\times$ \\
Training budget & $2.0\times 10^8$ tokens \\
Batch size & 2,048 (L0); 8,192 (L1) tokens \\
Optimizer & AdamW \\
Learning rate & $5\times 10^{-4}$ \\
Weight decay & 0 \\
Sparsity coefficients & $(12.0, 64.0)$ for layers 0 and 1 \\
Sparsity ramp & Linear ramp over full training budget \\
Sparsity scale & 4.0 \\
Dead-feature loss coefficient & $10^{-3}$ \\
Dead-feature window & $10^6$ tokens \\
JumpReLU backward bandwidth & 1.0 (L0); 0.10 (L1) \\
JumpReLU initial threshold & 0.03 \\
Decoder initialization norm & 0.1 \\
Initial target feature firing ratio & 0.10 \\
Normalization & Per-sample activation normalization \\
\hline
\end{tabular}
\end{table}

\clearpage
\section{Feature descriptors and token-baseline definitions}
\label{sec:supp_descriptors}

For feature $f$, participant $i$ and position $j$, let $a_{ijf}$ denote activation strength; $a_{ijf}>0$ defines an active position. Feature cards combine activation support, source diagnoses, temporal and demographic context, comorbidity, metadata, selected genotype annotations and output effects. Activation-mass weighting gives stronger firings greater weight. Source-token baselines ask whether the feature emphasizes a subset with different context from that expected merely because it activates on particular diagnoses.

For a categorical axis $z$ with bins $b$, the observed feature distribution was
\begin{equation}
P_f^z(b)=
\frac{
\sum_{i,j} a_{ijf}\mathbf{1}[z_{ij}=b]
}{
\sum_{i,j} a_{ijf}
}.
\end{equation}
For example, $z$ could be event age, calendar month, event year, ICD-10 chapter, event source, sex, or birth-year bin.

The token-level baselines were computed once from the full event table and did not depend on SAE features. For token $t$ and categorical axis $z$, the event-level token baseline was
\begin{equation}
B_t^z(b)=
\frac{
\sum_{i,j}\mathbf{1}[x_{ij}=t]\mathbf{1}[z_{ij}=b]
}{
\sum_{i,j}\mathbf{1}[x_{ij}=t]
}.
\end{equation}
For patient-level axes, such as sex or birth year, the analogous baseline used unique participants carrying token $t$ rather than event counts.

A feature-specific expected distribution was then obtained by mixing these fixed token baselines according to the feature's activation-mass distribution over source tokens:
\begin{equation}
\pi_{ft}=
\frac{
\sum_{i,j} a_{ijf}\mathbf{1}[x_{ij}=t]
}{
\sum_{i,j}a_{ijf}
},
\qquad
Q_f^z(b)=
\sum_t \pi_{ft} B_t^z(b).
\end{equation}
Thus $Q_f^z$ is feature dependent only through the feature's source-token mixture $\pi_{ft}$; the token baselines $B_t^z$ themselves are feature independent.

Categorical deviations were first summarized by the observed total variation distance,
\begin{equation}
\mathrm{TVD}_{\mathrm{obs},f}^z=
\frac{1}{2}\sum_b
\left|P_f^z(b)-Q_f^z(b)\right|.
\end{equation}
We report excess TVD after subtracting the expected discrepancy under an independent-draw binomial null. Activation-mass summaries use the Kish effective sample size. The correction limits finite-support inflation; it does not remove within-participant dependence or all confounding. The derivation and approximation are provided in Supplementary Section~\ref{sec:supp_tvd_detail}.

Continuous and count-valued card fields were summarized with activation-weighted means, token-matched expected means, and standardized differences. Context modules used the same principle: observed context-token rates around feature-active positions were compared with rates expected from the feature's source-token mixture, with direct same-token pairs excluded. Temporal context modules further stratified context rates by lag windows around the active event.

Patient-level feature summaries were computed by aggregating activations within a participant. These included patient mass Gini and the share of mass carried by the highest-mass participants, to flag features dominated by a small number of participants.

Baseline scope depends on the card module. Event-axis and diagnosis-context summaries use activating-token mixtures; some global patient-metadata annotations use cohort-wide reference distributions. A metadata association should therefore not be described as source-token-adjusted unless that field's reference supports the claim. Support and patient-mass concentration should be considered alongside effect size.

\section{Finite-sample correction for source-token enrichment}
\label{sec:supp_tvd_detail}

Because low-support features can have nonzero TVD even when their active positions are random draws from the expected distribution $Q_f^z$, we also computed an analytic finite-sample null. For an unweighted count-based summary with $n_f$ active observations and expected bin probability $q_b=Q_f^z(b)$, the null contribution of bin $b$ was
\begin{equation}
\frac{1}{2}\,
\mathbb{E}\left|
\frac{X_b}{n_f}-q_b
\right|,
\qquad
X_b \sim \mathrm{Binomial}(n_f,q_b).
\end{equation}
The expectation was evaluated analytically from the binomial marginal distribution. For activation-mass-weighted summaries, we used the same expression with the Kish effective sample size
\begin{equation}
n_{\mathrm{eff},f}=
\frac{\left(\sum_k w_k\right)^2}{\sum_k w_k^2},
\end{equation}
where $w_k$ is the activation mass contributing to the relevant feature-axis distribution. We then reported the excess TVD,
\begin{equation}
\mathrm{TVD}_{\mathrm{excess},f}^z
=
\mathrm{TVD}_{\mathrm{obs},f}^z
-
\mathbb{E}\!\left[\mathrm{TVD}_{\mathrm{null},f}^z\right],
\end{equation}
as the primary enrichment score in feature cards. The binomial expectation is exact under the assumed independent-draw null, not a correction for dependence between positions from the same participant. Activation-mass summaries use the effective-sample-size approximation.

\clearpage
\section{Genotype details}
\subsection{Selected genotype annotations}
\label{sec:supp_genotypes}

Beyond the mean dosage and its deviation from the token-mix baseline, we used genotype analyses to annotate features within a predefined set of genotypes, rather than to perform a genome-wide association study. For each SAE feature, we used the patient-level activation mass sum as the feature phenotype. For participant $i$ and feature $f$, let $m_{if}$ denote total patient activation mass, and let $g_{iv}$ denote allele dosage for variant $v$.

Feature phenotypes and genotype dosages were residualized against available covariates, including an intercept, sex, birth timing or age proxy, genotype principal components, and genotype batch indicators where available. Variants were filtered by missingness, minor-allele frequency, and genotype standard deviation. The primary association statistic was the Pearson correlation between residualized feature mass and residualized dosage:
\begin{equation}
r_{fv} =
\mathrm{corr}
\left(
\tilde{m}_{\cdot f},
\tilde{g}_{\cdot v}
\right).
\end{equation}
Approximate p-values were computed from these correlations, and Benjamini-Hochberg false-discovery rates were computed across the scanned feature-variant family ~\cite{benjamini1995controlling}.

For each feature we retained the top 50 variant hits by association p-value. The observed dosage was an activation-mass-weighted mean over feature-active positions; expected dosage mixed token-level dosage baselines using the same activation-mass token shares as the context analysis. Their difference describes excess dosage relative to that token mixture. The retained association rows and dosage contrasts are selected annotations, not the full scan's null distribution.

\subsection{Genotype scan quality control}
\label{sec:appendix_genotype_qc}

Of 12,887 input variants, 12,778 passed the scan filters. Missing dosages of retained variants were mean-imputed before residualization.
\begin{table}[pos=!ht]
\centering
\caption{Genotype scan filters and covariates.}
\label{tab:appendix_genotype_qc}
\small
\begin{tabular}{ll}
\hline
Setting & Value \\
\hline

Input variant panel & 12,887 SNPs (DisGeNET-linked to cohort disease set) \\

Variants retained for testing & 12,778 \\

Minimum observed samples per variant & 20 \\
Minor-allele-frequency filter & $\ge 0.005$ \\
Missingness filter & $\le 0.05$ \\
Genotype SD filter & $\ge 10^{-8}$ \\

Covariates & intercept, sex, birth year/month, genotype PCs, batch \\
Number of genotype PCs & 40 \\
Association statistic & Pearson $r$ on residualized mass and dosage \\
Multiple testing & Benjamini--Hochberg $q$ over feature--variant family \\
Hits retained per feature & top 50 by association $p$-value \\
\hline
\end{tabular}
\end{table}

\clearpage
\section{Feature atlas overview extras}

\begin{figure}[pos=!htbp]
\centering
\includegraphics[width=\linewidth]{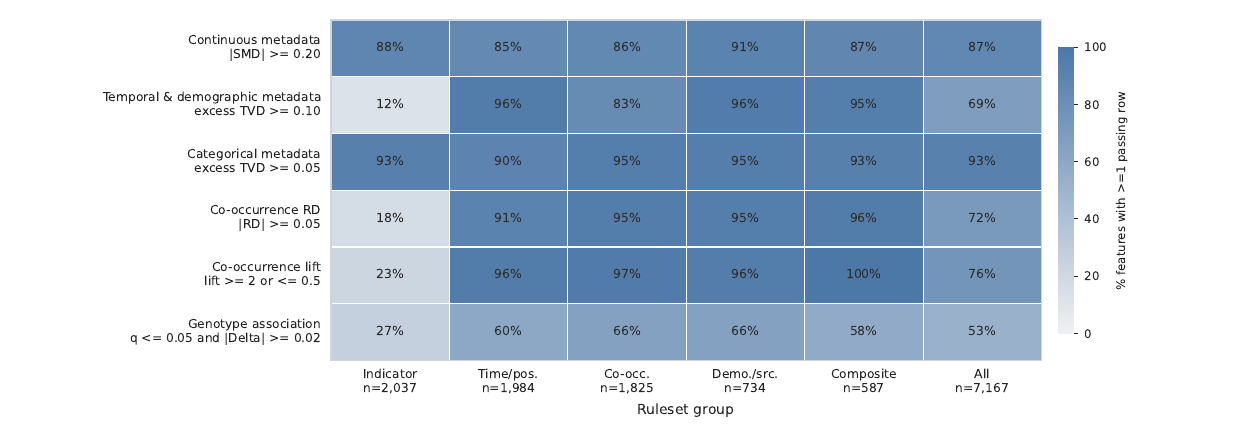}
\caption{\textbf{Features crossing card-display thresholds by ruleset group.} Each cell gives the percentage of features in its column with at least one statistic meeting the row's support and display criteria. Column denominators are the supported features in each rule-family group; ``All'' includes 7,167 features with at least 10 total activations. Thresholds are shown beside each row, with support and baseline definitions in Methods. Genotype entries additionally require $q\leq0.05$. These display summaries do not represent percentages of clinically validated features.}
\label{fig:features-above-card-gate}
\end{figure}

\begin{figure}[pos=!h]
    \centering
    
    \includegraphics[width=1\linewidth]{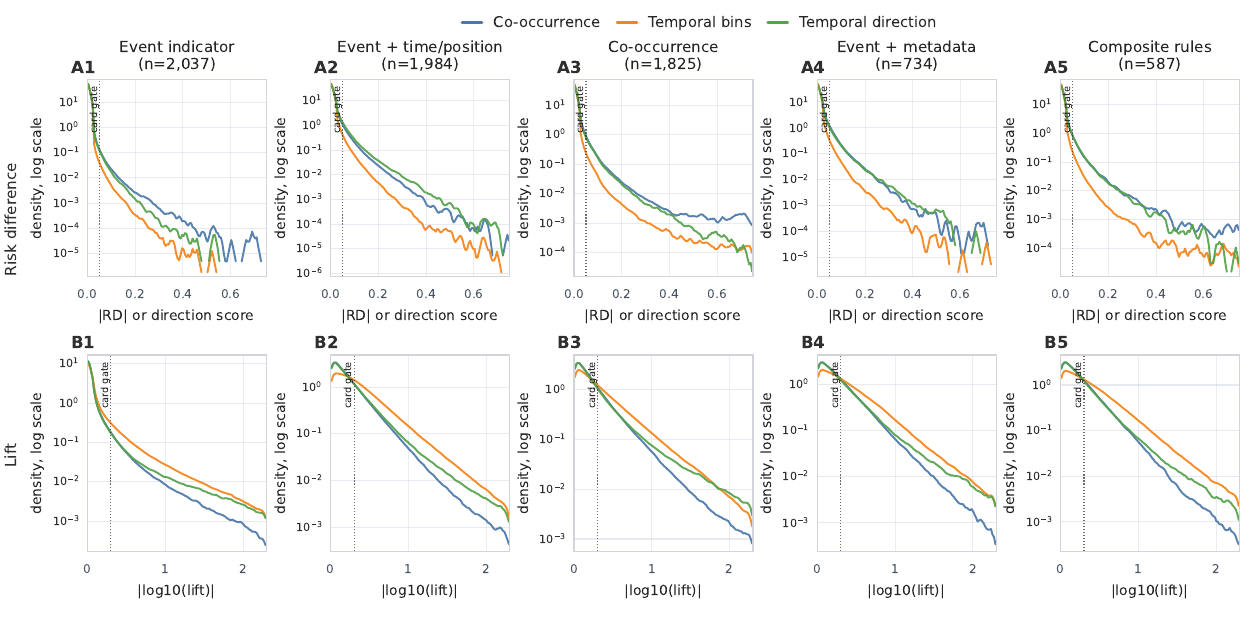}
    \caption{\textbf{Context risk-difference and lift distributions by ruleset family.}
Per-family expansion of the comorbidity and temporal context evidence in
Fig~\ref{fig:feature_atlas_overview}D1--D2, broken out by ruleset group (columns) and context
type (line color: co-occurrence, temporal bins, temporal direction).
(A1--A5) Absolute mass risk difference ($|$RD$|$) or temporal-direction score and
(B1--B5) absolute log10 mass lift, for the event-indicator, event + time/position,
co-occurrence, event + metadata, and composite-rule groups.
Row counts are shown on a log scale; curves are absolute row counts rather than normalized
densities, so height reflects the number of feature--context rows at a given effect size. Dotted
lines mark the card-display gate.}
\label{fig:atlas_context_appendix}
\end{figure}

\begin{figure}
    \centering

\includegraphics[width=1\linewidth]{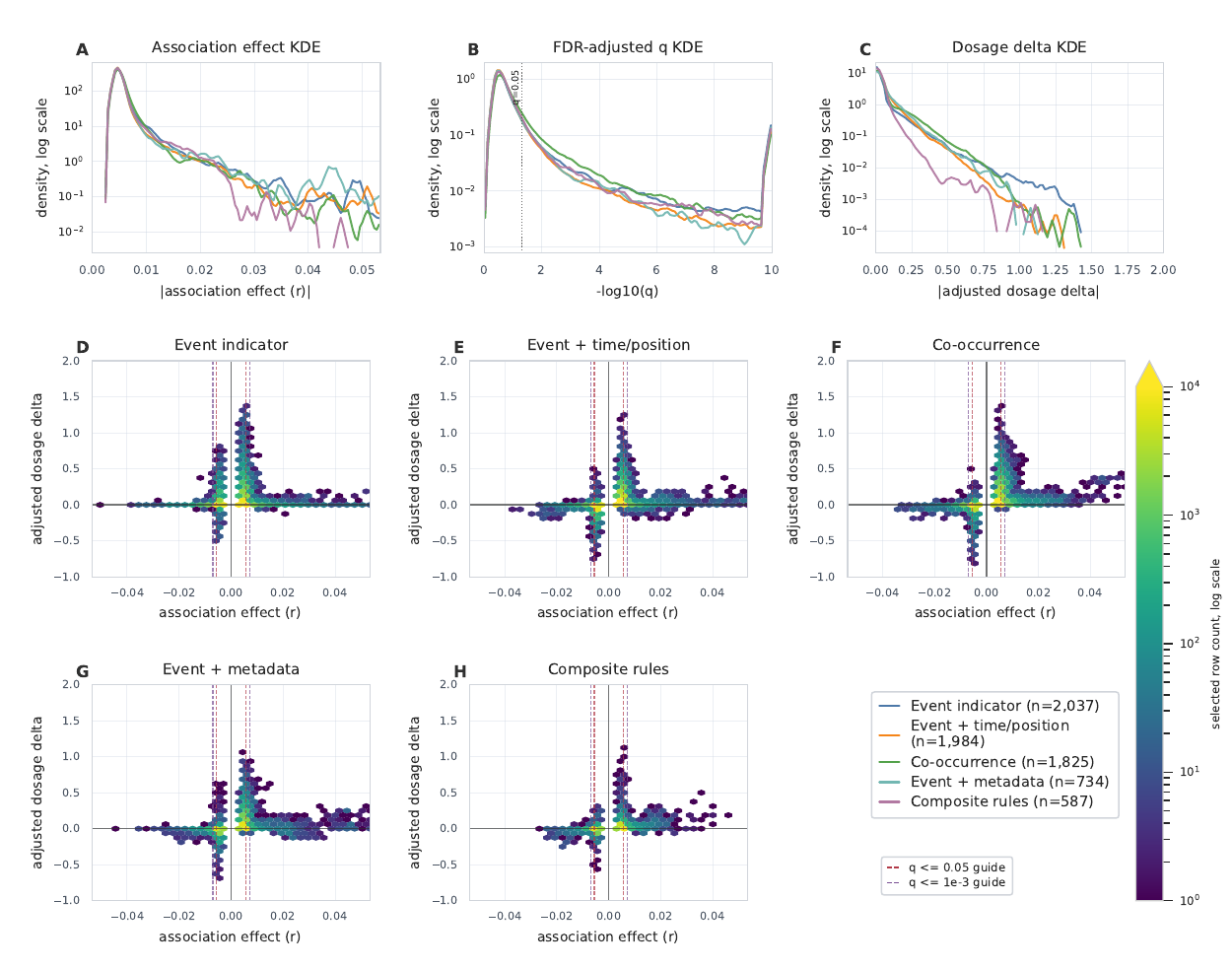}
    \caption{\textbf{Selected genotype-hit evidence by ruleset family.}
Per-family expansion of the genotype panel in Fig~\ref{fig:feature_atlas_overview}E, restricted
to selected feature--SNP hit rows (top 50 per feature).
(A) Kernel density of absolute association effect (residualized $r$),
(B) FDR-adjusted significance as $-\log_{10}(q)$ with a $q = 0.05$ guide, and
(C) absolute token-mix-adjusted dosage delta, each stratified by ruleset group (line color).
(D--H) Two-dimensional histograms of association effect ($r$) versus adjusted dosage delta for
the event-indicator, event + time/position, co-occurrence, event + metadata, and composite-rule
groups, colored by selected row count (log scale); dashed vertical guides mark empirical
effect-size thresholds corresponding to $q \le 0.05$ and $q \le 10^{-3}$.
All panels summarize selected top-hit rows and are not a full feature-by-SNP null distribution.}
\label{fig:atlas_genotype_appendix}
\end{figure}

\clearpage
\section{Ruleset induction details}
\label{sec:appendix_ruleset_config}

Candidate predicates were generated from the training split using the templates in Table~\ref{tab:appendix_predicate_templates}. Rules were induced by a beam search over predicate conjunctions followed by a greedy search over compact disjunctive-normal-form rules. Search used an activation-mass-aware composite heuristic. Its score $\mathrm{MCC}_{\mathrm{mass}}$ has the form of Eq.~\ref{eq:si_mcc}, but uses unnormalized positive activation mass and unit inactive-position weights. Atom scores combined $\mathrm{MCC}_{\mathrm{mass}}$ with 0.2 times ordinary MCC; candidate-rule scores combined $\mathrm{MCC}_{\mathrm{mass}}$, 0.1 times ordinary MCC and 0.05 times point-biserial correlation between rule predictions and activation magnitude. Both scores included a complexity penalty. Only the highest-scoring candidates were retained for expansion, and final rules were selected by the original activation-mass score on validation participants. The selected rules were then held fixed for evaluation using $\mathrm{MCC}_{\mathrm{SI}}$, with positive-activation normalization estimated from training participants only. Search settings are listed in Table~\ref{tab:ruleset-search-config}.

\begin{table}[pos=!ht]
\centering
\caption{Ruleset predicate templates.}
\label{tab:appendix_predicate_templates}
\small
\begin{tabular}{ll}
\hline
Predicate family & Template examples \\
\hline
Source event & token is $t$; token in top-$k$ active tokens; ICD chapter is $c$ \\
Event source & event-source categories from UK Biobank data-coding 2171 (e.g.\ self-report, primary care) \\
Calendar/date & month $=m$; day of week $=d$; year $\le/\ge c$ \\
Demographic & sex $=s$; birth year $\le/\ge c$; age $\le/\ge c$ \\
Trajectory position & position $\le/\ge c$; sequence length $\le/\ge c$ \\
Patient history & patient has token $t$; patient has no token $t$ \\
Temporal history & token $t$ occurs before position; token $t$ occurs after position \\
\hline
\end{tabular}
\end{table}

\begin{table}[pos=!ht]
\centering
\caption{\textbf{Ruleset search configuration.} Default search settings used for DNF ruleset mining. Candidate rules were generated on the training split, selected on validation data, and reported on the held-out test split.}
\label{tab:ruleset-search-config}
\begin{tabular}{ll}
\hline
Setting & Value \\
\hline
Candidate / selection / report splits & Train / validation / test \\
Minimum predicted support & 5 tokens \\
Maximum active/history/context tokens & 16 / 8 / 12 \\
Maximum temporal context tokens & 4 \\
Maximum atoms considered per conjunction & 48 \\
Maximum atoms per clause & 3 \\
Clause beam size & 64 \\
Maximum DNF clauses & 3 \\
Maximum saved candidates per feature & 80 \\
Complexity penalty & 0.006 per atom, plus extra clauses \\
\hline
\end{tabular}
\end{table}

\subsection{Scale-invariant evaluation and sensitivity}
\label{sec:supp_mcc_sensitivity}

Dividing positive weights by the feature's mean positive training activation preserves their relative strengths while removing dependence on uniform activation rescaling. The same training-derived divisor is used for held-out evaluation; no rule was reselected using test results. Search and selection retained the original activation-mass objective described above. Ordinary unweighted MCC provides a separate sensitivity analysis of binary occurrence alone.

Across the same 7,167 supported SAE features, median MCC was 0.725 with original activation weights, 0.699 with training-normalized weights and 0.627 without activation weighting (Table~\ref{tab:supp_mcc_sensitivity}). Original and normalized scores had Spearman correlation 0.985; 9.5\% of individual feature scores changed by more than 0.10. Event-indicator rules remained almost perfectly recoverable, and the broad confidence and recurrence associations were retained. Normalization affected random and dictionary baselines more strongly, motivating its use for cross-representation reporting. These comparisons hold rules and representation settings fixed; they do not establish that a different search objective would discover identical rules.

In a validation-only check of 7,056 features with identifiable selected candidates in the saved search outputs, switching from the original to the normalized metric displaced all best-original-score candidates for 16.5\% of features. The best normalized-score candidate improved normalized validation MCC over the best original-score candidate by more than 0.01 for 6.6\% and by more than 0.05 for 1.5\%. Ties among original-score winners were resolved by their highest normalized score. This check re-ranked stored candidates without rerunning discovery; the reported test results retain the original selected rules.

\begin{table}[pos=!htbp]
\centering
\caption{\textbf{Rule-evaluation sensitivity to activation weighting.} Held-out median MCC for fixed selected rules, using the same features with at least 10 total activations. The Original and Training-normalized columns use mass-weighted MCC with unnormalized and training-normalized positive weights, respectively. The latter defines the reported $\mathrm{MCC}_{\mathrm{SI}}$. Original weights were used for rule discovery and selection; ordinary MCC weights all positions equally. Representation sparsity budgets are unchanged.}
\label{tab:supp_mcc_sensitivity}
\small
\begin{tabular}{lrrr}
\hline
Representation & Original & Training-normalized & Ordinary \\
\hline
SAE & 0.725 & 0.699 & 0.627 \\
Thresholded hidden & 0.862 & 0.751 & 0.684 \\
Thresholded PCA & 0.646 & 0.636 & 0.495 \\
Random top-$k$ & 0.578 & 0.220 & 0.218 \\
Dictionary top-$k$ & 0.638 & 0.300 & 0.293 \\
\hline
\end{tabular}
\end{table}

\subsection{Language-model confidence and ruleset agreement}
\label{sec:supp_ruleset_summary_confidence}

\begin{figure}[pos=!htbp]
\centering
\includegraphics[width=\linewidth]{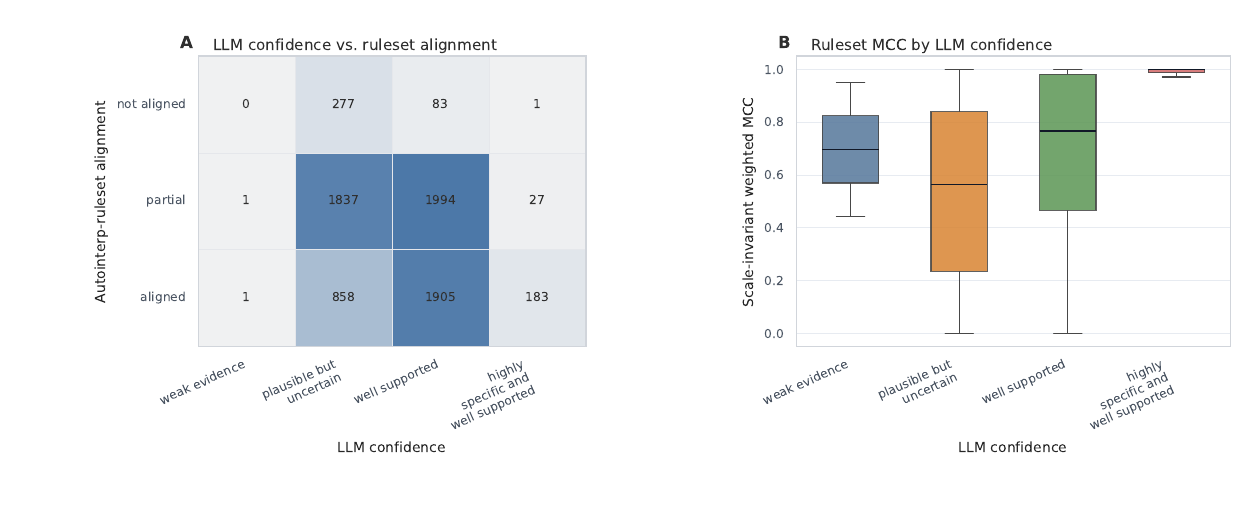}

\caption{\textbf{Language-model confidence, summary agreement and rule recoverability.} (A) Counts of features by self-rated summary confidence and summary--ruleset alignment. (B) Held-out mass-weighted MCC by confidence category. Counts refer to features, not participants. Boxes show medians and interquartile ranges; whiskers extend to observations within 1.5 interquartile ranges, with outliers omitted. Higher confidence is associated with greater agreement and recoverability, but is not a calibrated measure of clinical correctness.}
\label{fig:supp_ruleset_summary_confidence}
\end{figure}

The complementary summary checks from Section~\ref{sec:results_ruleset_checks} are shown in Fig.~\ref{fig:supp_ruleset_summary_confidence}. The confidence categories contained 2, 2,972, 3,982 and 211 features, respectively; the weak-evidence category is too small for a reliable group-level conclusion. Agreement with a fitted ruleset and higher rule recoverability do not independently establish the correctness of a clinical interpretation.

\section{Feature-stability matching score}
\label{sec:appendix_stability_score}
For a candidate feature pair $(f,g)$ within a layer, the hybrid similarity was
\begin{equation}
\begin{array}{rcl}
S(f,g)&=&0.30\,s_{\mathrm{dec}}(f,g)+0.25\,s_{\mathrm{act}}(f,g)+0.20\,s_{\mathrm{tok}}(f,g)\\
& &+\,0.10\,s_{\mathrm{ctx}}(f,g)+0.10\,s_{\mathrm{strat/lens}}(f,g)+0.05\,s_{\mathrm{rule}}(f,g),
\end{array}
\end{equation}
where $s_{\mathrm{dec}}$ is normalized decoder cosine similarity, $s_{\mathrm{act}}$ activation-overlap F1, $s_{\mathrm{tok}}$ top-token weighted Jaccard, $s_{\mathrm{ctx}}$ combined context/temporal similarity, $s_{\mathrm{strat/lens}}$ is agreement in demographic/calendar stratification summaries and logit-lens summaries, and $s_{\mathrm{rule}}$ ruleset agreement. A pair was a strong match when
\begin{equation}
S(f,g) \ge 0.70 \quad\mathrm{and}\quad \left(s_{\mathrm{act}}(f,g)>0 \ \mathrm{or}\ s_{\mathrm{tok}}(f,g)>0\right).
\end{equation}
Strong matches were aggregated across the three seeds into cross-run \emph{components} as the connected components of the strong-match graph, so a component groups feature-runs linked by one or more strong matches; it may contain a single seed's feature, one feature per seed, or occasional split/merge structures where one run's feature links to several in another. A component's \emph{recurrence} is the number of distinct seeds it spans (one, two, or three). The stability analyses were restricted to feature-runs with at least 100 activations (the activation-support gate), and reciprocal one-to-one pairs -- feature-runs that are mutual best matches between a pair of runs -- are reported separately as a clean subset for assessing match quality.

\subsection{Recurrence counts across the three SAE fits}

Table~\ref{tab:supp_stability_recurrence} separates feature-instance and component denominators. A feature instance is counted separately within each fitted atlas; a component groups matched instances across fits. Recurrent components contribute several instances, and occasional feature splits or merges mean that a component need not contain exactly one instance per fit. The two percentage summaries therefore answer different questions. All counts below use the support threshold of at least 100 activations. Raw patient-position identifiers were used only transiently to calculate activation overlap and were not written to the reliability outputs.

\begin{table}[pos=!htbp]
\centering
\caption{\textbf{Supported feature recurrence across three SAE fits.} Percentages use the column totals as denominators. The at-least-two-fits row combines the two- and three-fit rows. Instance counts include only support-passing features; component counts include components containing those instances.}
\label{tab:supp_stability_recurrence}
\small
\begin{tabular}{lrr}
\hline
Component recurrence & Feature instances, $n$ (\%) & Components, $n$ (\%) \\
\hline
One fit & 8,830 (47.0) & 8,830 (70.1) \\
Exactly two fits & 3,044 (16.2) & 1,533 (12.2) \\
All three fits & 6,928 (36.8) & 2,234 (17.7) \\
At least two fits & 9,972 (53.0) & 3,767 (29.9) \\
\hline
Total supported & 18,802 (100.0) & 12,597 (100.0) \\
\hline
\end{tabular}
\end{table}

\section{Masked steering: full dose responses and controls}
\label{sec:supp_masked_steering}

For each feature, 128 test participants with exactly one occurrence of the target diagnosis contributed a masked target event; 128 matched participants without that code in their available history contributed a masked other-diagnosis event. Controls were matched without replacement on generic diagnosis token type, age within five years, calendar year within five years and a sequence-length ratio of at most two, without selecting on activations or model scores. Chapter and block identifiers were cleared when masking. Code absence in the available history does not establish disease absence.

The same participants were evaluated at 81 signed doses from $-2$ to $+2$ in steps of 0.05, retaining the reconstruction residual and downstream normalization at the feature's native layer. The reference 75th percentiles of nonzero training activations were 1.1270 (L0F2228), 2.0083 (L1F68) and 2.3798 (L0F2979). All three features were inactive at the selected positions after masking, so negative doses subtract a fixed reference direction rather than ablate naturally active features. Zero-dose full-vocabulary logits agreed with the unchanged model to within $1.4\times10^{-5}$.

\begin{figure}[pos=!htbp]\centering\includegraphics[width=\linewidth]{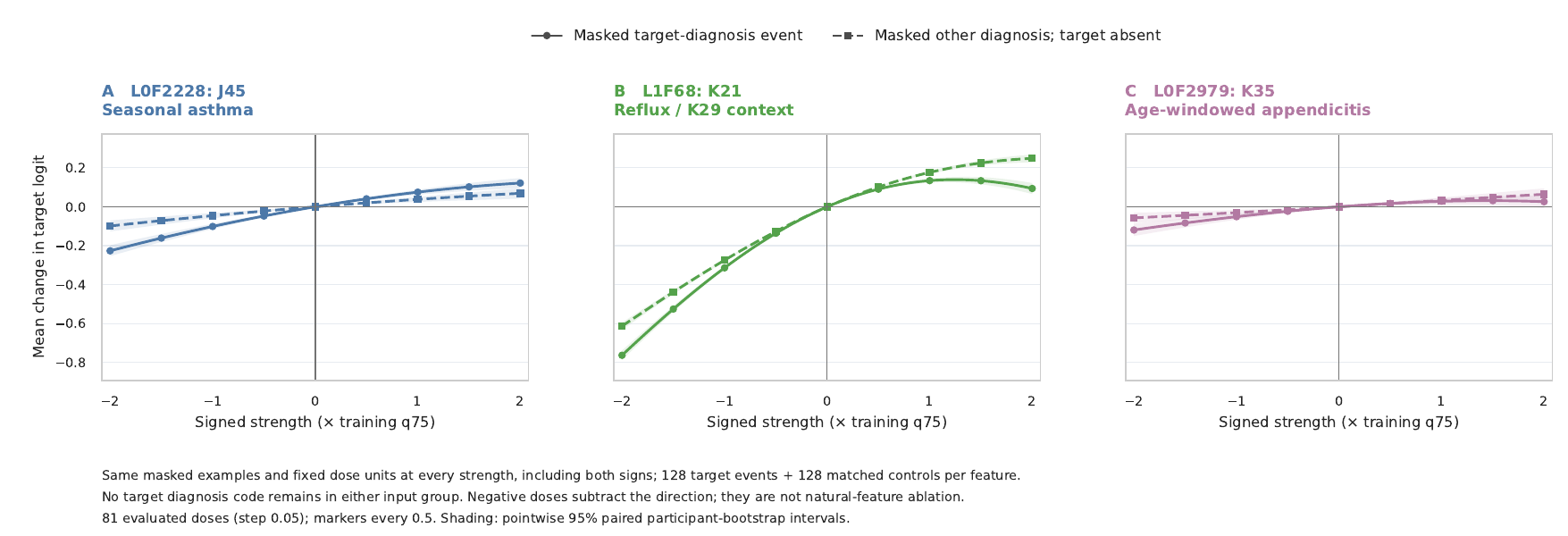}
\caption{\textbf{Full masked-steering dose responses and matched controls.} The three panels correspond to L0F2228 (J45, blue), L1F68 (K21, green), and L0F2979 (K35, purple). Solid curves show masked target events; dashed curves show masked other diagnoses from histories without the target code (128 participants per group and feature). All 81 signed doses from -2 to +2 in steps of 0.05 use identical examples and a common training-only q75 reference for both signs. Shading denotes pointwise 95\% paired participant-bootstrap intervals. Negative doses subtract a fixed decoder direction and are not natural-feature ablations. Both cohort-specific effects are retained rather than averaged. The main figure reports the -1 and +1 reference-dose target-event estimates.}
\label{fig:supp_masked_logit}\end{figure}

\begin{figure}[pos=!htbp]
\centering
\includegraphics[width=\linewidth]{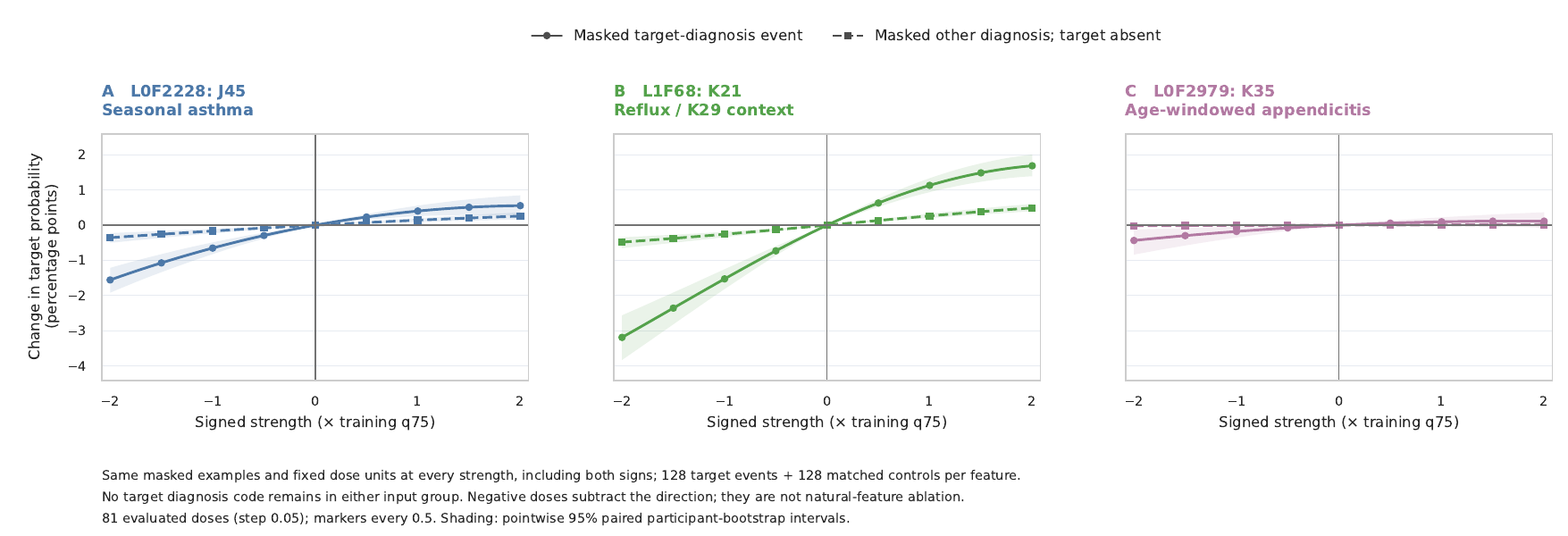}
\caption{Target-probability dose responses in percentage points, with matched controls and pointwise paired-bootstrap intervals.}
\label{fig:supp_masked_probability}
\end{figure}

\begin{figure}[pos=!htbp]
\centering
\includegraphics[width=\linewidth]{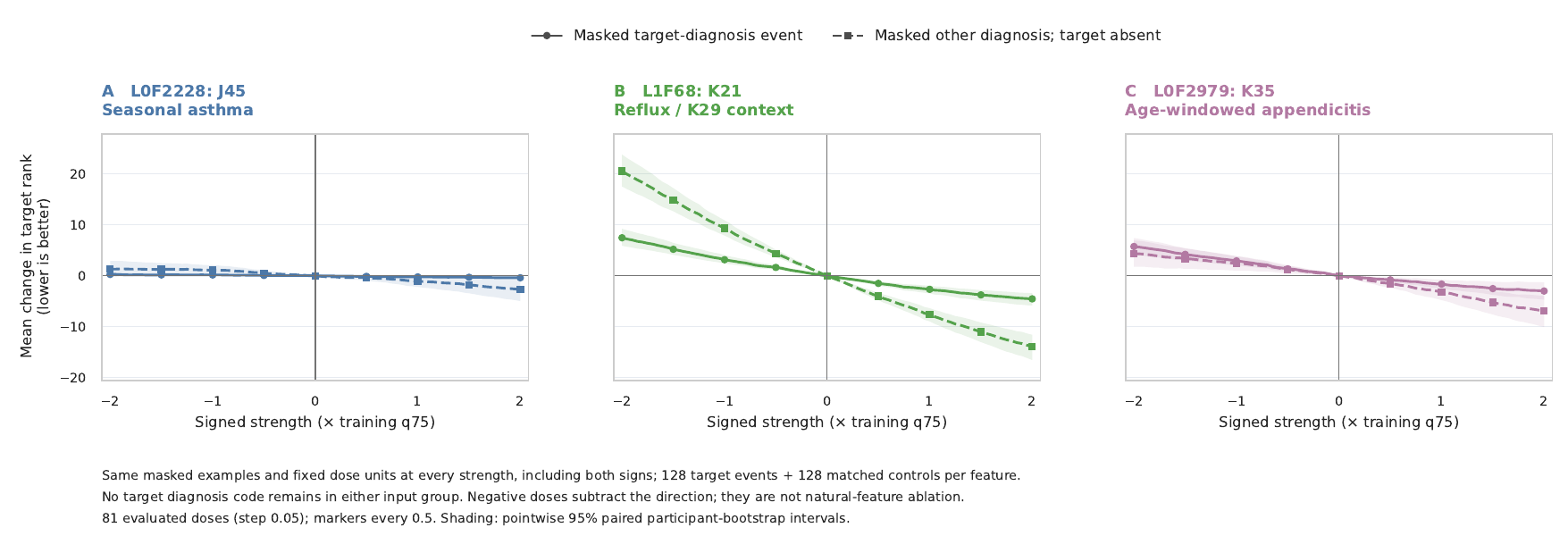}
\caption{Target-rank changes across signed doses. These complement logits and probabilities and do not establish diagnosis-specific benefit.}
\label{fig:supp_masked_rank}
\end{figure}

\begin{figure}[pos=!htbp]
\centering
\includegraphics[width=\linewidth]{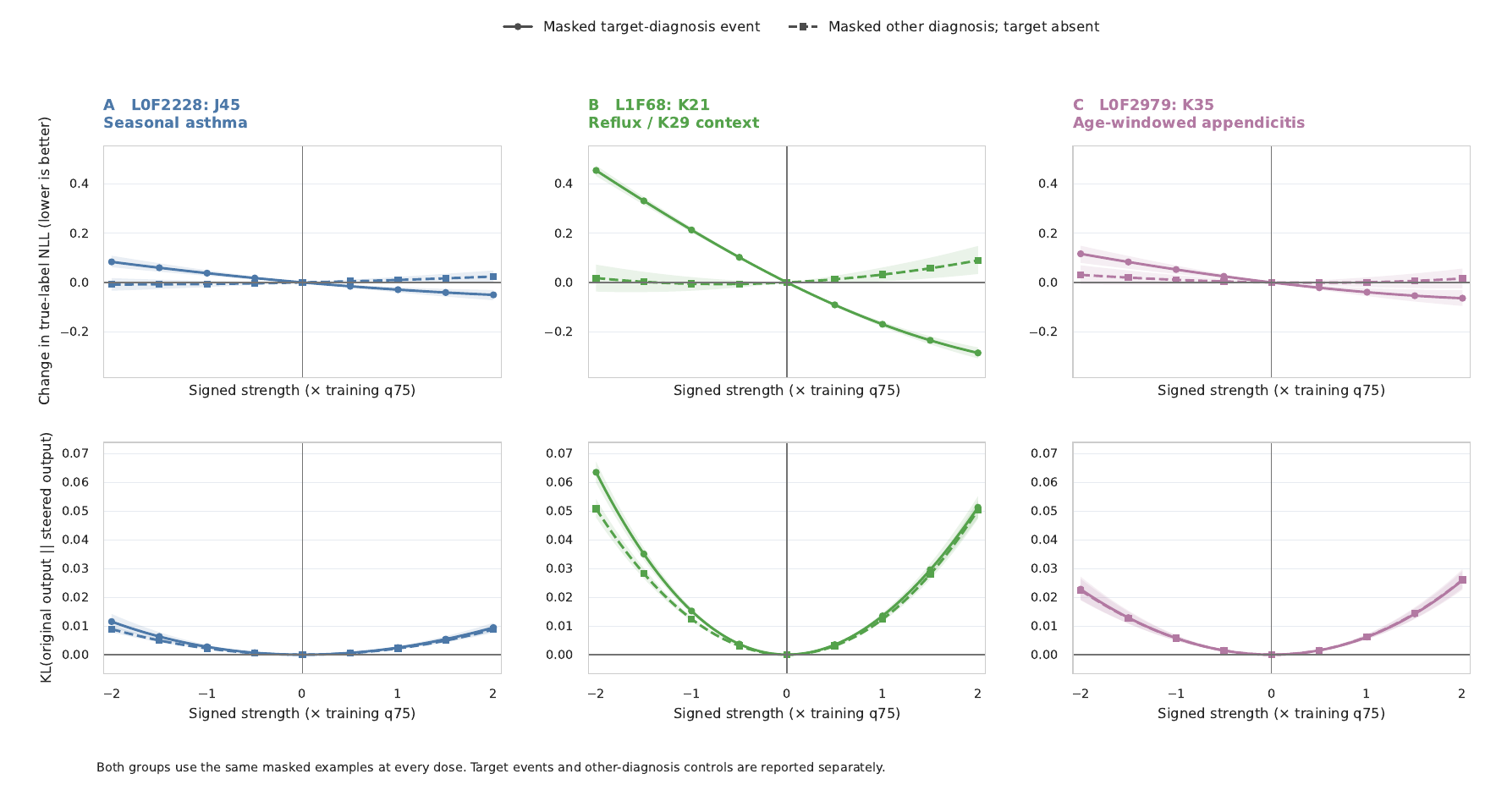}
\caption{Broader masked-output effects: target-independent prediction and distribution-change diagnostics on the same fixed cohorts.}
\label{fig:supp_masked_diagnostics}
\end{figure}

We evaluated target logits, probabilities and ranks, true-label negative log likelihood and full-output distribution changes (Figs.~\ref{fig:supp_masked_logit}--\ref{fig:supp_masked_diagnostics}). Probability and rank complement logits because changing one logit does not determine probability when other logits also change. Pointwise 95\% intervals used 5,000 paired participant-bootstrap resamples, fixed across doses within each cohort. These intervals describe participant sampling uncertainty conditional on the selected SAE direction, not simultaneous bands or uncertainty across SAE retrains. The main figure reports the $-1$ and $+1$ target-event estimates; the supplementary curves retain target and control cohorts separately.

\section{Sparse representation-baselines}
\subsection{Sparse representation-baseline construction}
\label{sec:appendix_baseline_construction}

For each layer and position $p$, let $h_p \in \mathbb{R}^D$ be the hidden state and $z_p$ a sparse baseline code. Thresholds or top-$k$ values were calibrated on the training split against the earlier SAE activation budget and retained after the layer-1 rerun. The final SAE averaged 18.3 active units per token-layer versus approximately 14 for the sparse comparators; these comparisons do not establish superiority at equal sparsity. For thresholded hidden and PCA baselines, coordinates were active when their absolute score exceeded a learned coordinate-specific threshold, $z_{pk}=s_{pk}\,\mathbf{1}\{|s_{pk}|\ge\tau_k\}$, where $s_{pk}$ is a hidden-state coordinate or PCA score; for random and dictionary baselines the largest-magnitude activations per position were retained. Reconstruction was summarized by FVU (Eq.~\ref{eq:fvu}); the dense hidden-state reference has $\mathrm{FVU}=0$ by definition.

\subsection{Representation-baseline comparison}
\label{sec:appendix_representation_baselines}

To contextualize the selected SAE atlas, we compared it with a dense hidden-state reference and four sparse representation baselines. The dense hidden-state reference corresponds to the original transformer residual stream and therefore has perfect reconstruction fidelity by definition, but it is not a sparse feature representation and was not rendered as feature cards. The sparse baselines were thresholded hidden-states, thresholded PCA directions, random top-$k$ directions, and learned dictionary top-$k$ features. The retained comparator settings use the earlier calibration described above. Features with fewer than 10 total activations were excluded before computing aggregate feature-card and ruleset statistics.

\begin{table}[pos=!ht]
\centering

\caption{{\bf Representation-baseline summary statistics.} Supported feature counts require at least 10 total activations. Of 8,192 SAE dictionary entries, 1,025 fall below this gate: 480 have no activations and 545 are live but have fewer than 10. Mass-weighted MCC uses training-normalized activation weights and is invariant to uniform feature rescaling. Boldface identifies the largest descriptive selectivity/recoverability statistic or lowest sparse-representation FVU; sparsity budgets differ.}
\label{tab:appendix_baseline_comparison}
\small
\setlength{\tabcolsep}{4pt}
\begin{tabular}{lrrrrrr}
\hline
Metric & Dense hidden & SAE & Hidden & PCA & Random & Dictionary \\
\hline
Feature-card interface & No & Yes & Yes & Yes & Yes & Yes \\

Supported feature count & -- & 7,167 & 1,024 & 1,024 & 3,236 & 8,184 \\

Mean active per token-layer & -- & 18.3 & 14.0 & 14.1 & 14.0 & 14.0 \\

Mean FVU & 0.000 & \textbf{0.002} & 0.809 & 0.786 & 0.672 & 0.075 \\
\noalign{\vskip 2pt}

Median TVD & -- & 0.214 & \textbf{0.467} & 0.126 & 0.435 & 0.414 \\

P90 TVD & -- & 0.630 & 0.755 & 0.223 & \textbf{0.824} & 0.767 \\
\noalign{\vskip 2pt}

Median context RD & -- & 0.156 & 0.131 & 0.059 & \textbf{0.205} & 0.186 \\

P90 context RD & -- & \textbf{0.674} & 0.202 & 0.112 & 0.460 & 0.436 \\
\noalign{\vskip 2pt}

Median mass-weighted MCC & -- & 0.699 & \textbf{0.751} & 0.636 & 0.220 & 0.300 \\
P90 mass-weighted MCC & -- & \textbf{1.000} & 0.913 & 0.828 & 0.576 & 0.764 \\
\hline
\end{tabular}
\end{table}


\section{Downstream task ablations}
\subsection{Prescription-derived TRD protocol and cohort}
\label{sec:supp_trd_protocol}

Treatment-resistant depression was operationalized from linked general-practice prescription issuance and diagnosis history. Drug name--quantity pairs were inner-joined to preferred PRESNER outputs~\cite{colon2023presner}; when preferred mapping rows duplicated a pair, the first stored preferred row was retained, while unmapped pairs were omitted. Candidate antidepressants were restricted to ATC class \texttt{N06A} and grouped by ChEMBL agent. Participants required an F32, F33 or F34 diagnosis and were excluded for recorded F31, F10--F19 or F20--F29 diagnoses. Issue dates for the same agent separated by no more than 14 weeks were joined into a course. Course start and end were the first and last observed issue dates, not inferred days supplied; courses spanning fewer than six weeks were not counted as adequate.

Adequate courses were ordered by start date. For a current course beginning at $s_c$ and the preceding course ending at $e_p$, a switch required a different agent, no overlap (including exclusion of shared end/start dates) and a directional gap
\begin{equation}
0 < s_c-e_p \leq 14\ \mathrm{weeks}.
\end{equation}

The positive proxy required at least two qualifying switches, implying at least three adequate courses. Course-gap and overlap logic were checked with regression tests. Participants without an adequate course were excluded. The label represents prescribing patterns and inferred treatment sequences, not adherence, dose adequacy, clinical response or adjudicated resistance.

The fixed assignments contained 693/10,436, 91/1,489 and 184/2,980 positive/eligible participants in training, validation and test, respectively. Outcome-specific prevalence was 6.6\%, 6.1\% and 6.2\%. All analyses below use these labels.

\begin{table}[pos=!htbp]
\centering
\caption{Prescription-derived TRD cohort by split.}
\label{tab:supp_trd_counts}
\begin{tabular}{lrrr}
\toprule
Split & Eligible & Positive & Prevalence \\
\midrule
Train & 10,436 & 693 & 6.6\% \\
Validation & 1,489 & 91 & 6.1\% \\
Test & 2,980 & 184 & 6.2\% \\
All & 14,905 & 968 & 6.5\% \\
\bottomrule
\end{tabular}
\end{table}

\begin{table}[pos=!htbp]
\centering
\caption{Disclosure-controlled demographics by corrected outcome. Percentages are within outcome. Five- and ten-year birth bands did not pass complementary-cell control, so the privacy gate selected 15-year bands.}
\label{tab:supp_trd_demographics}
\small
\begin{tabular}{>{\raggedright\arraybackslash}p{\dimexpr0.16\linewidth-2\tabcolsep\relax}>{\raggedright\arraybackslash}p{\dimexpr0.28\linewidth-2\tabcolsep\relax}>{\raggedleft\arraybackslash}p{\dimexpr0.28\linewidth-2\tabcolsep\relax}>{\raggedleft\arraybackslash}p{\dimexpr0.28\linewidth-2\tabcolsep\relax}}
\toprule
Panel & Category & TRD-negative, $n$ (\%) & TRD-positive, $n$ (\%) \\
\midrule
Sex & Female & 9,823 (70.5\%) & 704 (72.7\%) \\
 & Male/unknown & 4,114 (29.5\%) & 264 (27.3\%) \\
Birth cohort & 1935--1949 & 5,943 (42.6\%) & 411 (42.5\%) \\
 & 1950--1964 & 6,820 (48.9\%) & 490 (50.6\%) \\
 & 1965--1979 & 1,174 (8.4\%) & 67 (6.9\%) \\
\bottomrule
\end{tabular}
\end{table}

\begin{table}[pos=!htbp]
\centering
\caption{Overall adequate-course characteristics of the eligible cohort. These bins describe the operational phenotype construction and are not stratified by outcome because the switch threshold defines that outcome.}
\label{tab:supp_trd_courses}
\small
\begin{tabular}{llr}
\toprule
Measure & Category & Participants, $n$ (\%) \\
\midrule
Qualifying switches & 0 & 11,890 (79.8\%) \\
 & 1 & 2,047 (13.7\%) \\
 & 2+ & 968 (6.5\%) \\
Adequate courses & 1 & 4,054 (27.2\%) \\
 & 2 & 2,969 (19.9\%) \\
 & 3 & 2,064 (13.8\%) \\
 & 4+ & 5,818 (39.0\%) \\
\bottomrule
\end{tabular}
\end{table}

PRESNER mapping coverage could not be recovered from the fixed split and corrected-label artifacts. Estimating it would require a new row-level join to raw general-practice prescriptions; that join was not performed for this table. All displayed cells and complementary cells contain at least 10 participants.


\subsection{TRD readout configuration and performance}
\label{sec:supp_trd_models}

\label{sec:appendix_downstream_config}

The primary decoder-folded readout used all 32,768 pooled SAE inputs through a 2,048-dimensional reconstructed-hidden logistic fit, without an SAE-input support filter or log transform. The supplementary direct-SAE fit retained inputs active in at least 20 training participants, selected at most 2,048 by training activation mass and applied $\log(1+x)$. These pipeline differences mean their comparison is not an isolated test of decoder folding. Diagnosis bag-of-words (BOW) used binary presence of tokens observed in at least 20 training participants, together with sex and one-hot birth-year-bin covariates; no separate age covariate was included in that comparator.

The engineered comparator appended 111 outcome-agnostic variables to diagnosis BOW and demographics: 27 diagnosis-burden, age/calendar-span, gap, entropy and missingness summaries; 14 event-source count/fraction variables; 32 ICD-10-chapter count/fraction variables; 20 fixed age-bin variables; and 18 fixed calendar-bin variables. Summaries used all exported first-occurrence diagnosis rows in the eligible cohort; recorded-event span was not treated as continuous observation duration. Source and chapter levels present in at least 20 training participants were retained. Age cut points were 0, 10, \ldots, 80 years and calendar cut points were 1950, 1960, \ldots, 2020, with underflow and overflow bins; both counts and fractions were included. Non-finite ages and years were omitted from finite-value summaries and bins. When no finite value was available, the corresponding summary was set to zero and explicit missing-age and missing-date fractions retained the missingness information; missing event sources formed a separate candidate level. Prescription identities, issue dates, course counts and switch variables were excluded from every predictor matrix because they define the outcome. 

Linear readouts used L2-regularized, class-balanced logistic loss with training-fitted scaling that did not center sparse matrices. The inverse-regularization grid was
\[
C\in\{1,0.3,0.1,0.03,0.01,0.003,0.001,3\!\times\!10^{-4},10^{-4},3\!\times\!10^{-5},10^{-5}\}.
\]
Validation AUPR selected one candidate and test metrics were computed after model selection. The end-to-end MLP concatenated final-layer \texttt{BOS}, DOB and \texttt{EOS} states and used two 256-unit ReLU layers. The bidirectional-GRU comparator concatenated the same anchors with a bidirectional GRU summary of event states. AdamW used body/head learning rates $2\times10^{-5}$/$10^{-3}$; default (dropout 0.1, weight decay 0.01, three epochs) and stronger-regularization (dropout 0.3, weight decay 0.05, eight epochs) candidates were compared by validation AUPR with patience two and seed zero. Because the outcome was uncommon, we report average precision (reported as AUPR), AUPR divided by test prevalence, and area under the receiver-operating-characteristic curve (AUROC).

Performance intervals and model differences used 5,000 paired bootstrap resamples of the same 2,980 test participants. Coefficient intervals used 200 training-participant bootstrap refits and are conditional on the selected SAE and regularization.

\begin{table}[pos=!htbp]\centering\small
\caption{\textbf{TRD discrimination.} The same 2,980 test participants include 184 positives. Intervals condition on fitted models and the fixed split. Input transforms and selection differ across the implemented pipelines.}
\label{tab:trd_corrected_models}
\begin{tabular}{>{\raggedright\arraybackslash}p{\dimexpr0.44\linewidth-2\tabcolsep\relax}>{\raggedleft\arraybackslash}p{\dimexpr0.28\linewidth-2\tabcolsep\relax}>{\raggedleft\arraybackslash}p{\dimexpr0.28\linewidth-2\tabcolsep\relax}}
\toprule
Model & AUPR (95\% CI) & AUROC (95\% CI) \\
\midrule
Demographics & 0.069 [0.057, 0.083] & 0.536 [0.493, 0.579] \\
Demographics + BOW & 0.081 [0.063, 0.108] & 0.551 [0.507, 0.595] \\
engineered clinical history & 0.111 [0.085, 0.154] & 0.625 [0.585, 0.666] \\

Direct SAE (sensitivity) & 0.119 [0.087, 0.162] & 0.632 [0.593, 0.670] \\
SAE decoder-folded & 0.124 [0.093, 0.170] & 0.643 [0.605, 0.682] \\
Hidden linear & 0.127 [0.097, 0.174] & 0.655 [0.615, 0.695] \\
Finetuned BiGRU & 0.135 [0.104, 0.182] & 0.678 [0.640, 0.717] \\
\bottomrule\end{tabular}\end{table}

\begin{table}[pos=!htbp]
\centering\small
\caption{\textbf{Paired test differences from engineered clinical history.} Values are candidate minus comparator, with 95\% intervals from 5,000 paired resamples of the same 2,980 test participants. Intervals are not multiplicity-adjusted; exclusion of zero does not establish clinical utility.}
\label{tab:supp_trd_paired_differences}
\begin{tabular}{lrr}
\toprule
Candidate & $\Delta$AUPR [95\% CI] & $\Delta$AUROC [95\% CI] \\
\midrule
SAE decoder-folded & $+0.0134$ [$-0.0127$, $+0.0401$] & $+0.0187$ [$-0.0037$, $+0.0423$] \\
Hidden linear & $+0.0158$ [$-0.0070$, $+0.0423$] & $+0.0303$ [$+0.0091$, $+0.0519$] \\
Finetuned BiGRU & $+0.0237$ [$-0.0043$, $+0.0550$] & $+0.0537$ [$+0.0222$, $+0.0859$] \\
\bottomrule
\end{tabular}
\end{table}

\subsection{Decoder-folded readout and contribution screening}
\label{sec:appendix_trd_decoder_fold}

For each token position $p$, the SAE reconstructs normalized residual coordinates, $\widehat{\widetilde h}_p=W_Dz_p+b_D$. The TRD readout uses their mean across clinical-event positions and their values at BOS, EOS and DOB; it does not restore each token's raw hidden-state norm before pooling. Thus, event-mean pooling commutes with affine decoding:
\begin{equation}
\frac{1}{|\mathcal E|}\sum_{p\in\mathcal E}\widehat{\widetilde h}_p
=W_D\left(\frac{1}{|\mathcal E|}\sum_{p\in\mathcal E}z_p\right)+b_D,
\end{equation}
where $\mathcal E$ denotes a participant's clinical-event positions. For each layer and pooling block, absorb the fitted classifier-input scaling into $q$ and the intercept $b$. Writing $\widehat{\widetilde h}$ and $z$ for the corresponding pooled vectors gives
\begin{equation}
\eta=b+q^\top\widehat{\widetilde h}=(b+q^\top b_D)+(W_D^\top q)^\top z.
\end{equation}
Thus $\gamma=W_D^\top q$ gives exact SAE coefficients for that block, with block contributions summed for the full readout. Coordinate-wise maximum pooling generally does not commute with affine decoding, and the token-specific normalization $s(h_p)$ must not be interchanged with pooling. The folded and reconstructed-hidden logits agreed across all 14,905 eligible participants to maximum absolute error $9.6\times10^{-6}$ (mean $1.3\times10^{-6}$). This verifies equivalence to the classifier on normalized SAE reconstructions; it does not eliminate reconstruction error relative to the original hidden states or assume that the MLM's final normalization is affine.

A standardized coefficient $\gamma_js_{j,\mathrm{train}}$ is the model's formal logit slope per training standard deviation with other coordinates held fixed. The normalization removes arbitrary activation units, but the hypothetical independent coordinate change may not be realizable through the SAE encoder. It is not a causal change in patient risk. Dependence is not unique to SAEs; decoder folding additionally restricts the allowable coefficient vectors to the column space of $W_D^\top$.

For descriptive interpretation, we ranked supported inputs by absolute bootstrap-median standardized coefficient and grouped dependent inputs before screening. A training activation graph connected inputs with absolute Pearson correlation at least 0.80 and at least 10 coactive training participants; connected components defined disjoint groups. Connected components can contain transitively linked members that are not themselves strongly pairwise correlated; the group is interpreted from its full membership. Named inputs required at least 20 active training and 10 active validation participants. Groups were ordered by the highest-ranked member, termed the anchor. The anchor required at least 95\% coefficient-sign consistency, a 5th--95th empirical interval excluding zero, and agreement between point and median signs. In addition, at least 90\% of bootstrap refits had to preserve the group's training contribution pattern at correlation at least 0.80 with its point-fit pattern.

For each observed participant, the centered group contribution was $c_G=\sum_{j\in G}\gamma_j(z_j-\bar z_{j,\mathrm{train}})$. We required $\operatorname{Var}(c_G)/\sum_{j\in G}\operatorname{Var}(c_j)\geq0.25$ and $\operatorname{Cov}(c_G,\eta)/\operatorname{Var}(c_G)\geq0.50$ in both training and validation. The second quantity measures net alignment with the complete score, including all remaining readout inputs; it is a descriptive regression slope and may exceed one or be negative. The first ten qualifying groups were retained. The graph contained 5,756 supported nonconstant inputs, whereas the complete-score calculation used all 32,768 inputs. These exploratory criteria select stable patterns with limited net offset; they do not establish feature independence or unique predictive importance. Selection used validation data previously used for readout tuning and was not an independent confirmatory analysis. Nearby thresholds were evaluated as sensitivity checks.

Writing $c_{-G}$ for all remaining terms, net score alignment can also be written as
\begin{equation}
K_G=\frac{\operatorname{Cov}(c_G,\eta)}{\operatorname{Var}(c_G)}
=1+\frac{\operatorname{Cov}(c_G,c_{-G})}{\operatorname{Var}(c_G)}.
\end{equation}
It uses the full score, not just pairwise cancellation within selected pairs. However, it is a distribution-level linear summary over observed participants: it neither proves cancellation for every patient nor rules out hidden states outside that distribution. Multiplying a coefficient by $K_G$ would not produce a new independent or causal coefficient. Group removal without refitting decomposes the current score; held-out removal/refitting experiments would be needed for a different claim about unique predictive performance.

\begin{table}[pos=!htbp]
\centering\small
\caption{\textbf{Complete membership of the ten displayed groups.} Groups are numbered in the order of the displayed inputs in Table~\ref{tab:trd_screened_sae_groups}. Each displayed input is its group\textquotesingle s highest-ranked member, termed the anchor. Connected components permit groups with more than two inputs. Event denotes mean clinical-event pooling; BOS, EOS and DOB denote sequence-start, sequence-end and birth-date positions. The ten groups contain 30 distinct pooled inputs.}
\label{tab:supp_group_members}
\begin{tabular}{>{\raggedleft\arraybackslash}p{\dimexpr0.075\linewidth-2\tabcolsep\relax}>{\raggedright\arraybackslash}p{\dimexpr0.235\linewidth-2\tabcolsep\relax}>{\raggedright\arraybackslash}p{\dimexpr0.69\linewidth-2\tabcolsep\relax}}
\toprule
Rank & Anchor & Complete pooled-input membership \\
\midrule
1 & L0F1080-event & L0F1080-event \\
2 & L0F3798-event & L0F3798-event, L1F2470-event, L1F1250-event \\
3 & L1F2043-event & L1F2043-event, L1F2043-EOS \\
4 & L1F2182-event & L1F2182-event, L1F2182-EOS, L1F3173-event, L1F3027-event, L0F246-event, L1F2182-BOS, L1F2182-DOB \\
5 & L0F1200-event & L0F1200-event \\
6 & L0F832-event & L0F832-event \\
7 & L0F2118-DOB & L0F2118-DOB, L1F1879-DOB \\
8 & L1F3442-EOS & L1F3442-EOS, L1F2312-BOS, L1F1721-event, L1F965-event, L0F1360-event, L1F2730-event \\
9 & L1F4079-EOS & L1F4079-EOS \\
10 & L0F3284-event & L0F3284-event, L1F332-EOS, L1F771-BOS, L1F3467-event, L1F1458-event, L1F2676-event \\
\bottomrule
\end{tabular}
\end{table}

\begin{table}[pos=!htbp]
\centering\small
\caption{\textbf{Coefficient uncertainty and contribution-screening results.} Group numbers follow Supplementary Table~\ref{tab:supp_group_members}. The coefficient interval is the anchor's 5th--95th percentile across 200 refits. $R_{\min}$ is the smaller training/validation value of $\operatorname{Var}(c_G)/\sum_{j\in G}\operatorname{Var}(c_j)$; $K_{\min}$ is the smaller value of net score alignment $K_G$. Refit agreement is the percentage of refits whose training group-contribution correlation with the point fit was at least 0.80. Primary thresholds were $R_{\min}\geq0.25$, $K_{\min}\geq0.50$ and agreement at least 90\%. These summarize observed contribution patterns, not independent effects.}
\label{tab:supp_trd_group_diagnostics}
\begin{tabular}{rrrrr}
\toprule
Rank & Anchor coeff. [5th, 95th] & $R_{\min}$ & $K_{\min}$ & Refit agreement (\%) \\
\midrule
1 & $+0.072$ [$+0.032$, $+0.109$] & 1.000 & 5.400 & 99.5 \\
2 & $+0.069$ [$+0.048$, $+0.092$] & 1.315 & 1.030 & 100.0 \\
3 & $-0.057$ [$-0.081$, $-0.031$] & 0.939 & 2.527 & 96.0 \\
4 & $+0.054$ [$+0.024$, $+0.083$] & 0.925 & 4.718 & 91.0 \\
5 & $+0.050$ [$+0.025$, $+0.072$] & 1.000 & 2.077 & 100.0 \\
6 & $-0.047$ [$-0.080$, $-0.017$] & 1.000 & 6.766 & 99.5 \\
7 & $+0.039$ [$+0.008$, $+0.074$] & 1.883 & 7.715 & 98.0 \\
8 & $+0.037$ [$+0.018$, $+0.055$] & 1.587 & 2.774 & 92.0 \\
9 & $+0.034$ [$+0.023$, $+0.046$] & 1.000 & 3.496 & 100.0 \\
10 & $+0.030$ [$+0.015$, $+0.043$] & 4.000 & 1.596 & 100.0 \\
\bottomrule
\end{tabular}
\end{table}

Nearby threshold checks retained all ten anchors except when the required fraction of agreeing refits changed from 0.90 to 0.80 or 0.95; each variant then retained eight primary anchors (Table~\ref{tab:supp_screen_sensitivity}). Changing the activation-correlation threshold altered group membership while retaining the anchors. These one-at-a-time checks describe sensitivity to the tested settings, not exhaustive robustness or independent validation.

\begin{table}[pos=!htbp]
\centering\small
\caption{\textbf{Sensitivity of the selected list to screening thresholds.} Each variant selects ten groups and changes one primary setting. Shared anchors counts overlap with the primary ten; member Jaccard is intersection divided by union of the complete selected pooled-input sets.}
\label{tab:supp_screen_sensitivity}
\begin{tabular}{>{\raggedright\arraybackslash}p{\dimexpr0.60\linewidth-2\tabcolsep\relax}>{\raggedleft\arraybackslash}p{\dimexpr0.20\linewidth-2\tabcolsep\relax}>{\raggedleft\arraybackslash}p{\dimexpr0.20\linewidth-2\tabcolsep\relax}}
\toprule
Setting & Shared anchors & Member Jaccard \\
\midrule
Primary settings & 10 & 1.000 \\
Activation-correlation threshold = 0.75 & 10 & 0.857 \\
Activation-correlation threshold = 0.85 & 10 & 0.900 \\
Coefficient-sign consistency = 0.9 & 10 & 1.000 \\
Coefficient-sign consistency = 0.975 & 10 & 1.000 \\
Minimum within-group variance ratio = 0.1 & 10 & 1.000 \\
Minimum within-group variance ratio = 0.5 & 10 & 1.000 \\
Minimum net score alignment = 0.25 & 10 & 1.000 \\
Minimum net score alignment = 0.75 & 10 & 1.000 \\
Minimum fraction of agreeing refits = 0.8 & 8 & 0.590 \\
Minimum fraction of agreeing refits = 0.95 & 8 & 0.472 \\
\bottomrule
\end{tabular}
\end{table}

\printcredits

\newpage

\bibliographystyle{cas-model2-names}

\bibliography{trail-refs}



\end{document}